%% file: main.tex
\documentclass[Afour,sageh,times]{sagej}

\usepackage{graphicx}%
\graphicspath{{figures/}{plots/}}%
\usepackage{multirow}%
\usepackage{amsmath,amssymb,amsfonts}%
\usepackage{mathrsfs}%
\usepackage[title]{appendix}%
\usepackage{etoolbox}
\patchcmd{\appendices}{\quad}{: }{}{}
\usepackage{xcolor}%
\usepackage{textcomp}%
\usepackage{manyfoot}%
\usepackage{booktabs}%
\usepackage{listings}%
\usepackage{subfig}%
\usepackage{siunitx}%
\usepackage{enumitem}
\usepackage{mdframed}
\usepackage{mathtools}
\usepackage{algorithm}
\usepackage[noEnd,indLines]{algpseudocodex}
\usepackage{subdepth}
\usepackage{multicol}
\usepackage{float}

\let\emptyset\varnothing

\algrenewcommand\algorithmicrequire{\textbf{Input:}}
\algrenewcommand\algorithmicensure{\textbf{Output:}}

\newlength{\continueindent}
\makeatletter
\newcommand*{\ALG@customparshape}{\parshape 2 \leftmargin \linewidth \dimexpr\ALG@tlm+\continueindent\relax \dimexpr\linewidth+\leftmargin-\ALG@tlm-\continueindent\relax}
\apptocmd{\ALG@beginblock}{\ALG@customparshape}{}{\errmessage{failed to patch}}
\makeatother

\usepackage[colorlinks,bookmarksopen,bookmarksnumbered,citecolor=red,urlcolor=red]{hyperref}
\usepackage{cleveref}

\def\volumeyear{2024}

\begin{document}

\runninghead{Piedra \textit{et~al.}}

\title{Visuotactile and Explicitly Force-Controlled Robotic Ultrasound for Abdominal Volumetric Reconstruction
}

\author{Adrian Piedra\affilnum{1}, R Brooke Jeffrey\affilnum{2}, and Oussama Khatib\affilnum{1}}

\affiliation{\affilnum{1}Stanford Robotics Laboratory, Computer Science Department, Stanford University, Stanford, CA 94305 USA\\
\affilnum{2}Department of Radiology, School of Medicine, Stanford University, Stanford, CA 94305 USA}

\corrauth{Adrian Piedra,
Stanford Robotics Laboratory,
Stanford University,
Stanford, CA
94305 USA}

\email{apiedra@stanford.edu}

\begin{abstract}
In this paper, we present a robotic ultrasound acquisition system that integrates stereo vision, touch-based feedback, and expert-informed strategies to perform autonomous and adaptive abdominal scans. The system records freehand motion and force data from expert radiologists, creating a framework to capture transducer motion, applied forces, and anatomical scanning strategies. This expert data is replayed to replicate characteristic scans with the robot, forming a foundation for further autonomous capabilities. Using stereo vision, the system generates three-dimensional topography maps of the patient's abdomen, which are refined through stiffness measurements at key points to delineate the rib cage boundary. These combined techniques enable the robot to execute two distinct scanning paths: an upward-angled sweep beneath the rib cage to visualize structures near the upper abdomen and a perpendicular sweep across soft tissue regions.
A compliant, torque-controlled seven degree-of-freedom robotic manipulator is controlled to maintain consistent probe contact through closed-loop force control over the varied anatomical surfaces.
Physical experiments demonstrate that the system achieves high-quality imaging comparable to expert scans while dynamically adapting to patient-specific topographies.
Furthermore, the robotic system surpasses expert capabilities by enabling three-dimensional volume acquisition, which enhances diagnostic potential and provides volumetric data for advanced analyses.
This work highlights the integration of expert knowledge into autonomous robotic systems and underscores the potential of combining perception-based autonomy with physical reasoning for enhanced diagnostic performance.
\end{abstract}

\keywords{Robotic ultrasound, autonomy, expert strategy, stereo-based topography, rib boundary delineation}

\maketitle

\input{sections/1-introduction}
\input{sections/2-related_work}
\input{sections/3-framework}

\input{sections/4-enabling_autonomy}
\input{sections/5-experimental_results}

\input{sections/6-conclusion}

\begin{acks}
    The authors would like to thank the Stanford Ultrasound Imaging and Instrumentation Lab, and in particular Andrew Andrzejek, for providing access to the ImFusion software that enabled the three-dimensional ultrasound volume reconstruction and segmentation.
\end{acks}















\end{document}

%% file: sections/1-introduction.tex
\section{Introduction}\label{sec:intro}
\begin{figure*}[t]%
    \centering
    \includegraphics[width=\linewidth]{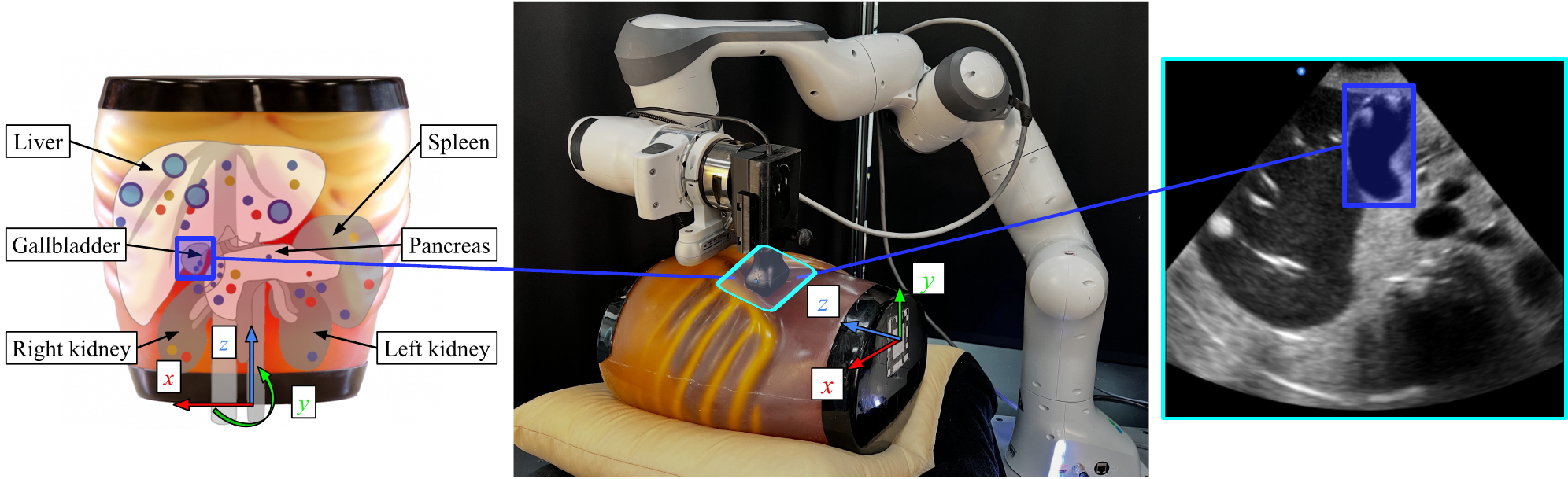}
    \caption{
    Robotic ultrasound acquisition system (with a focus on the gallbladder as an example).
    \textbf{Left}: Abdominal ultrasound phantom with RAS coordinate frame and labeled pathologies.
    \textbf{Middle}: Seven
    degree-of-freedom
    robot manipulating ultrasound transducer over the gallbladder.
    \textbf{Right}: Robotically acquired ultrasound image with the gallbladder highlighted.}
    \label{fig:system}
\end{figure*}
Ultrasound is the most widely used diagnostic imaging modality in the world today. Its major advantages compared to other forms of imaging include its low cost, unparalleled safety due to the lack of ionizing radiation, and ease of portability. As a result, ultrasound is a critical tool in medical settings ranging from hospitals to remote field clinics.
Ultrasound plays a vital role in providing timely and accurate diagnoses, enabling medical professionals to assess conditions such as cardiac disease, liver pathology, and prenatal health without the need for complex and costly equipment. However, despite its advantages, ultrasound has one major drawback that has limited its continued growth and accessibility: it requires highly-trained sonographers to acquire the images with a hand-held transducer.

The need for skilled ultrasound technologists has created a significant bottleneck in healthcare delivery. There is currently a world-wide shortage of trained ultrasound professionals, and this shortage is particularly severe in low-resource areas where access to healthcare services is already limited. Compounding this issue is the fact that ultrasound technicians are often prone to ergonomic injuries due to the repetitive motions involved in operating the transducer. These challenges have led to an increasing need for
solutions that can
improve access to high-quality diagnostic imaging.

One promising solution is the use of robotic systems to assist
human technologists in performing ultrasound scans. Robotic ultrasound systems can house the ultrasound transducer on a robotic manipulator that moves and adjusts according to controller inputs, allowing for the acquisition of a general survey exam. These systems have the potential to dramatically reduce the burden on human operators by performing an initial survey scan autonomously, and in turn, address both the shortage of trained professionals and the ergonomic challenges faced by human technicians.


Robotic ultrasound systems equipped with external vision and force sensors are capable of moving toward autonomous capabilities. These systems can not only plan and perform the scan, but also analyze the resulting images and provide initial diagnostic insights. For example, artificial intelligence algorithms can assist in the real-time interpretation of ultrasound images, highlighting areas of concern or confirming the presence of abnormal findings.

The deployment of robotic ultrasound systems also enables
remote operators, such as skilled ultrasound physicians, to control the system from afar through haptic feedback technology. These remote physicians can guide the transducer's movements, adjust its angle, and apply the necessary force on the patient's body to optimize image quality. This remote operation is particularly valuable in rural or underserved areas, where access to specialized medical expertise may be limited. Through teleoperation, skilled physicians can remotely supervise and adjust scans.

The implications of robotic ultrasound systems are profound. They can increase access to diagnostic imaging in areas with limited access to healthcare professionals, reducing wait times and potentially saving lives. In addition, these technologies can be deployed in emergency settings, mobile clinics, and during disasters, where the ability to provide immediate medical imaging is critical for patient outcomes. Furthermore, the growing potential for autonomy in these systems can enhance diagnostic accuracy and the overall efficiency of healthcare delivery.

This paper is organized as follows.~\Cref{sec:related} provides an overview of literature related to the use of robotics in the field of ultrasound imaging.
\Cref{sec:framework} introduces a framework for robotic ultrasound acquisition, including a method for recording and analyzing a radiologist's scanning strategies and an implementation of a unified force-motion robot controller for ultrasound, with the robotic system shown in~\Cref{fig:system}.
\Cref{sec:autonomy} details the integration of vision and touch to enable the robot's autonomous generation and execution of scanning paths.
\Cref{sec:results} presents experimental results on ultrasound scans acquired (1) directly by a radiologist, (2) by a robot referencing the radiologist's freehand strategies, and (3) autonomously by a robot using vision and touch. Finally,~\Cref{sec:conclusion} concludes this paper.

%% file: sections/2-related_work.tex
\section{Related Work}\label{sec:related}

The field of robotic ultrasound imaging has seen significant advancements in recent years, driven by the integration of robotics, artificial intelligence, and advanced imaging technologies. These developments address critical limitations in manual ultrasound procedures, including operator fatigue, variability in image quality, and restricted access to medical treatment, especially in remote or underserved areas. Robotic ultrasound systems are classified into teleoperated and autonomous systems, with each category addressing specific challenges and capabilities.

\subsection{Teleoperated Robotic Ultrasound Systems}
Teleoperated systems have played a pivotal role in expanding access to diagnostic ultrasound imaging, particularly in scenarios where the physical presence of a clinician is not feasible.~\cite{Chatelain2015} demonstrated how robotic systems can improve ultrasound imaging by ensuring consistent image quality and reducing operator dependency.

In such systems, a remote operator directly controls the ultrasound transducer.~\cite{Conti2014} introduced a teleoperation system that allows human operators to perform imaging procedures from significant distances.~\cite{Jorda2022} extended this concept by enabling real-time haptic-robotic interactions for ultrasound procedures across distances exceeding $\SI{10000}{\kilo\meter}$, using haptic feedback to replicate the tactile experience of a physical examination.

Such systems rely heavily on operator skill, making them suitable for highly skilled clinicians but challenging to scale in contexts with a shortage of trained personnel. Studies by~\cite{Fu2023} and~\cite{Si2024} show that advances in teleoperation interfaces, such as haptic-feedback devices, augmented reality overlays, and predictive force models, have further improved usability and accuracy in remote ultrasound procedures.

\subsection{Autonomous Robotic Ultrasound Systems}
Autonomous robotic ultrasound systems aim to reduce dependency on human operators by enabling robots to perform imaging tasks independently.
\cite{Jiang2023} reviewed the progression of robotic ultrasound systems, emphasizing the importance of modeling radiologists' decision-making processes to replicate expert-level semantic reasoning.~\cite{Li2021} outlined the need for fully autonomous systems capable of scan path planning, image interpretation, and real-time adaptation to patient anatomy. Their review identified critical challenges in force control, image quality optimization, and patient-specific adaptations.

Artificial intelligence (AI) and machine learning (ML) are at the core of autonomous robotic ultrasound systems. These methods improve diagnostic accuracy by enhancing image quality, identifying anatomical landmarks, and optimizing scan trajectories.~\cite{Huang2017} demonstrated the use of reinforcement learning for robotic systems to adapt dynamically to patient-specific variations.
Other studies, such as those by~\cite{Machado2018} and~\cite{Bi2024}, have explored the role of supervised and unsupervised learning models in detecting and classifying abnormalities like tumors, cysts, and vascular anomalies.




While vision-based methods
enable initial scan path planning, inaccuracies in visual sensing necessitate tactile feedback for real-time trajectory refinement.
Physical autonomy ensures that robotic systems can adapt to patient-specific anatomical variations, dynamically adjust probe pressure, and maintain consistent imaging quality.~\cite{Bamaarouf2024} demonstrated the importance of compliant control strategies for achieving safe and effective interactions between the robot and patient.
Systems like the UltraGelBot presented by~\cite{Raina2024}, which autonomously applies ultrasound gel and adapts to variations in patient anatomy, exemplify the growing emphasis on touch-based methods in enabling autonomous robotic ultrasound.





\cite{Santos2018} presented a hierarchical control architecture for robotic-assisted tele-echography, combining explicit Cartesian force control with joint torque-based orientation control. The system uses a three-dimensional time-of-flight camera and a force sensor to anticipate contact stiffness before engagement, adapting control gains dynamically. This predictive approach minimizes jerky movements during transitions and ensures smooth and stable interactions with the patient's anatomy. In contact, a stiffness estimation algorithm leverages the relationship between perceived stiffness and the robot's effective mass at the end-effector, allowing the system to maintain force tracking in unstructured environments.


\subsection{Integration of Teleoperated and Autonomous Systems}
The synergy between teleoperated and autonomous systems has emerged as a promising solution for future robotic ultrasound imaging. Autonomous systems can perform initial survey scans, acting as a first-pass diagnostic filter to determine if a patient appears healthy. If abnormalities are detected or the system encounters uncertainties, the teleoperation scheme connects an expert to guide further imaging as though the clinician were physically present with the patient.
\cite{Kuo2023} introduced a hybrid robotic ultrasound system combining autonomous contact force regulation and teleoperated control, demonstrating clinical viability in phantom-based experiments.~\cite{Fu2021} emphasized the role of compliant robotic manipulators in enabling seamless transitions between autonomous and teleoperated modes, which helps to maintain patient safety and diagnostic quality.

In our previous work, we introduced a robotic ultrasound imaging framework that utilized a haptic interface to enable long-distance remote operation and presented a basic robotic replay of expert freehand scanning strategies~\citep{Piedra2024}. This replay approach relied on recorded expert probe motion, applied force, and narrated observations to replicate freehand scanning actions with a robot. While effective in reproducing the expert's scanning strategy, the system did not include advanced autonomous capabilities for independent scan planning or rib boundary delineation.


\subsection{Three-Dimensional Volume Acquisition}
Three-dimensional (3D) volume acquisition offers significant advantages over traditional two-dimensional (2D) ultrasound imaging by providing volumetric data that allows clinicians to explore entire regions of interest. Unlike 2D imaging, which captures specific planes, 3D ultrasound enables reconstruction of any plane post-acquisition, reducing operator dependence and improving diagnostic accuracy. This advantage is especially beneficial for AI and ML models, as the richer spatial context of 3D data enhances feature extraction and reduces errors due to inconsistent acquisition angles or missing anatomical details~\citep{Duffy2024}. Furthermore, novice-acquired 3D volumes have been shown to yield diagnostic results comparable to expert-acquired 2D scans, highlighting the utility of 3D acquisition in reducing operator dependency~\citep{Salinaro2021}.

Robotic systems are uniquely suited for generating 3D ultrasound volumes due to their programmable velocity and smooth, jerk-free motion, ensuring consistent scan paths and reducing variability between scans. This precision enables robots to repeat portions of a scan with high fidelity, improving reproducibility and data quality~\citep{Bharadwaj2022}. Additionally, robotic systems can maintain constant probe contact with adaptive force control, enhancing resolution and minimizing noise in 3D volumes. These capabilities improve lesion segmentation and quantitative measurements, such as fetal biometry or organ volume estimation, demonstrating the need for 3D ultrasound in clinical diagnostics~\citep{Valente2022}.

\subsection{Diagnostic Applications and Future Directions}
Autonomous robotic ultrasound systems have demonstrated utility across diverse clinical applications, including cardiac, musculoskeletal, and obstetric imaging.~\cite{Tang2024} highlighted the role of autonomous systems in cardiac imaging, where precise transducer control is essential for visualizing dynamic structures like the heart. Additionally, immersive platforms integrating augmented reality and virtual reality like the one in~\cite{Shyam2023} have been proposed to enhance operator understanding and control, offering new paradigms for semi-autonomous systems.

Future challenges include integrating multimodal sensing, achieving regulatory approval, and conducting large-scale clinical trials to validate system efficacy. Continued research will likely focus on combining AI/ML with touch-based control for comprehensive physical and image autonomy, such as the work presented by~\cite{Haxthausen2021}.

The literature underscores the ongoing challenges and future directions in the field of robotics in ultrasound imaging. Although the potential benefits are evident, there are still technical hurdles, cost considerations, and regulatory barriers that need to be addressed. Additionally, ensuring patient safety and comfort during autonomous scans requires more careful consideration. As technology advances, the integration of robotics in ultrasound imaging is expected to grow,
potentially expanding access to healthcare services in remote underserved areas and improving overall patient care.

%% file: sections/3-framework.tex
\section{Framework for Robotic Ultrasound}\label{sec:framework}
We present a robotic ultrasound acquisition system composed of a compliant, torque-controlled seven degree-of-freedom
robot manipulator and ultrasound transducer. These components and the ultrasound phantom, along with its labeled anatomy and RAS anatomical coordinate frame,
are shown in~\Cref{fig:system}.



\begin{figure*}[t]%
    \centering
    \includegraphics[width=\linewidth]{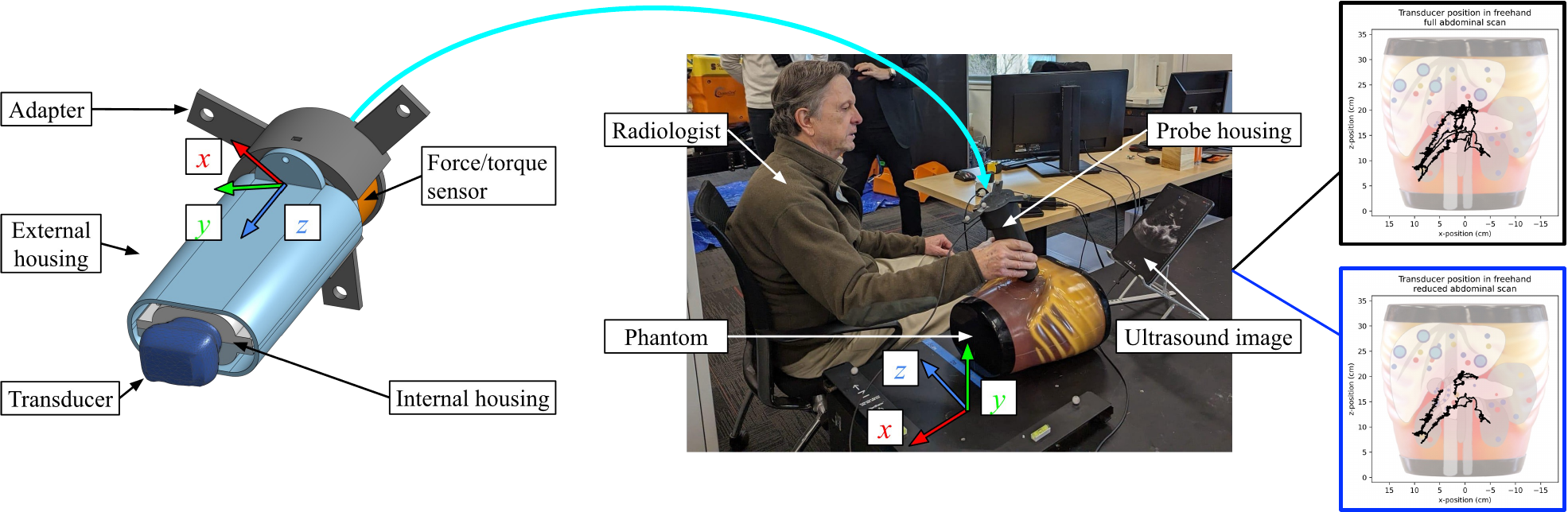}
    \caption{Freehand ultrasound scan setup. \textbf{Left}: Mechanism for recording ultrasound transducer position and orientation, along with the contact force between the transducer and phantom. \textbf{Middle}: Freehand scan recording setup (motion capture system frame labeled). \textbf{Top right}: Recorded freehand motion over phantom surface. \textbf{Bottom right}: Trimmed freehand motion without redundancy.}
    \label{fig:freehand_scan}
\end{figure*}

\subsection{Operational Space Robot Controller}\label{sec:robot_controller}
We use the unified motion and force controller in operational space to locally control both the motion and force behavior on the robot~\citep{Khatib1987}. With this decoupling technique and assuming perfect estimates of the dynamic components, the motion and force control loops regulate the desired behavior of the decoupled end-effector equivalent to a unit mass system in motion space and a force source in force space. This unified control strategy provides good performance in controlling both the motion of the transducer and its contact with the phantom.

Consider a robot end-effector in contact with the environment.
The robot's joint space dynamics are given by
\begin{align}
    A(q) \ddot{q} + b(q, \dot{q}) + g(q) + J_c^T F_c = \Gamma \label{eq:joint_space},
\end{align}
where $q$ is the joint space coordinates, $A$ is the robot mass matrix, $b$ is the vector of centrifugal and Coriolis forces, $g$ is the joint gravity vector, $J_c$ is the contact Jacobian, $F_c$ is the vector of contact forces, and $\Gamma$ is the joint torque vector. We choose the operational space task point to be the contact point such that the task Jacobian is equal to the contact Jacobian, $J = J_c$. After multiplying~\eqref{eq:joint_space} by the transpose of the dynamically consistent inverse of the task Jacobian, $\bar{J}^T = \Lambda J A^{-1}$, the task space dynamics become:
\begin{align}
    \Lambda \ddot{x} + \mu + p + F_c = F \label{eq:op_space},
\end{align}
where $\Lambda$ is the task space inertia matrix, $x$ is the task space coordinates, $\mu$ is the vector of task space centrifugal and Coriolis forces, $p$ is the task space gravity vector, $F$ is the task force, and we have
\begin{align}
    \Lambda &= \left( J A^{-1} J^T \right)^{-1}\mkern-3.5mu\relax, \\
    \mu &= \bar{J}^T b - \Lambda \dot{J} \dot{q}, \\
    p &= \bar{J}^T g.
\end{align}

The end-effector's translation and rotation are $x_t$ and $x_r$. The end-effector's instantaneous linear and angular velocity are $v$ and $\omega$. These quantities are represented by
\begin{align}
    x &= \begin{bmatrix}
        x_t \\ x_r
    \end{bmatrix},
    & \dot{x} &= \begin{bmatrix}
        v \\ \omega
    \end{bmatrix}.
\end{align}
The task forces and moments are denoted by $F$. In this paper, we represent the end-effector's rotation with the matrix $R \in \mathbb{R}^{3 \times 3}$. The end-effector rotation in task space, $x_r$, is derived from $R$ according to the selected task representation.

When the robot is in contact, its workspace is separated into the force space (directions of constraints) and motion space (directions of free motion). In translation, the force and motion spaces are defined by orthogonal projection matrices $\Omega_{f,t}$ and $\Omega_{m,t}$, respectively, such that $\Omega_{f,t} + \Omega_{m,t} = I_{3 \times 3}$. Similarly, in rotation, the force and motion spaces are defined by orthogonal projection matrices $\Omega_{f,r}$ and $\Omega_{m,r}$, respectively, such that $\Omega_{f,r} + \Omega_{m,r} = I_{3 \times 3}$. The resulting six-dimensional force and motion space projections are given by the following block diagonal matrices, hereafter referred to as the task specification matrices:
\begin{align}
    \Omega_f &= \begin{bmatrix}
        \Omega_{f,t} & 0_{3 \times 3} \\
        0_{3 \times 3} & \Omega_{f,r}
    \end{bmatrix},
    & \Omega_m &= \begin{bmatrix}
        \Omega_{m,t} & 0_{3 \times 3} \\
        0_{3 \times 3} & \Omega_{m,r}
    \end{bmatrix}.
\end{align}


Let the desired end-effector position and rotation in task coordinates be denoted $x_{d,t}$ and $x_{d,r}$, respectively.
The desired end-effector rotation matrix, $R_d$, is used to derive
$x_{d,r}$.
The instantaneous angular error corresponding to the error between the actual end-effector rotation and its desired rotation is expressed as
\begin{align}
    \delta \phi = E_r^+(x_r - x_{d,r}),
\end{align}
where the matrix $E_r$ maps the basic Jacobian to the task Jacobian, $J = E_r J_0$, as a function of the task representation.
The end-effector error in motion is then
\begin{align}
    e_m = \begin{bmatrix}
        e_{m,t} \\
        e_{m,r}
    \end{bmatrix}
    = \begin{bmatrix}
        x_t - x_{d,t} \\
        \delta \phi
    \end{bmatrix}.
\end{align}

The control force in motion space, $F_m$, is generated by an operational space proportional-derivative controller given by
\begin{align}
    F_m = \hat{\Lambda} \Omega_m
    \Bigl( -K_{p,m}
    e_m
    - K_{v,m}
    \dot{x}
    \Bigr), \label{eq:robot_controller_motion}
\end{align}
The estimate of the task space inertia matrix, $\hat{\Lambda}$, is used to dynamically decouple the system. The proportional and derivative gains for motion control are $K_{p,m}$ and $K_{v,m}$.

The control force in force space, $F_f$, is computed to follow the desired end-effector force, $F_d$, through a proportional-integral controller with feedforward force compensation and velocity-based damping as follows:
\begin{align}
\begin{split}
    F_f = \Omega_f \biggl( F_d &- K_{p,f} (F_s - F_d) \\
    &- K_{i,f} \int (F_s - F_d) - K_{v,f}
    \dot{x}
    \biggr). \label{eq:robot_controller_force}
\end{split}
\end{align}
The force space proportional, derivative, and integral gains are $K_{p,f}$, $K_{v,f}$, and $K_{i,f}$. The sensed task force, $F_s$, is used as feedback to the controller.

The motion and force controllers are composed together with two additional gravity and Coriolis/centrifugal compensation terms. The result is the dynamically decoupled and unified motion and force control force, $F$, given by
\begin{align}
    F = F_f + F_m + \hat{\mu} + \hat{p},
\end{align}
where $\hat{\mu}$ is the estimated vector of task space Coriolis and centrifugal forces, and $\hat{p}$ is the estimated task space gravity.

\subsubsection{Controlling Redundancy.}
Since the robot manipulator is redundant with respect to the task of controlling the transducer six-dimensional motion and force, the redundancy is controlled with a joint configuration-holding control torque vector $\Gamma_0$ projected onto the dynamically consistent task null-space, $N$.
The joint space control torques are
\begin{align}
    \Gamma = J^T F + N^T \Gamma_0,
\end{align}
where $N = I_{7 \times 7} - \bar{J}J$.


\subsubsection{Selecting the Task Specification Matrices.}\label{sec:task_specification}
Rather than controlling local moments applied to the phantom,
the robot end-effector's rotation is always controlled with pure motion control, where $\Omega_{m,r} = I_{3 \times 3}$ and $\Omega_{f,r} = 0_{3 \times 3}$.

Multiple contact directions are uncommon in abdominal ultrasound scans, especially with the reduction in friction forces due to the ultrasound gel applied between the transducer and abdomen. Therefore, when the end-effector is in contact with the abdomen in the direction denoted by the unit vector $\hat{n} \in \mathbb{R}^3$, the force space projection matrix in translation is
\begin{align}
    \Omega_{f,t} &= \hat{n} \hat{n}^T, \label{eq:force_space_contact}
\end{align}
and the motion space projection matrix in translation is $\Omega_{m,t} = I_{3 \times 3} - \hat{n} \hat{n}^T$.
This gives the following task specification matrices, with the matrix sizes removed, as
\begin{align}
    \Omega_f &= \begin{bmatrix}
        \hat{n} \hat{n}^T & 0 \\
        0 & 0
    \end{bmatrix},
    & \Omega_m &= \begin{bmatrix}
        I - \hat{n} \hat{n}^T & 0 \\
        0 & I
    \end{bmatrix}. \label{eq:task_specification}
\end{align}


In free space, the robot's task specification matrices are set for pure motion control in translation and rotation, with $\Omega_{m} = I_{6 \times 6}$ and $\Omega_{f} = 0_{6 \times 6}$.
This is equivalent to setting the contact direction in~\eqref{eq:force_space_contact} as the zero vector,
\begin{align}
    \hat{n} \coloneqq 0_{3 \times 1}. \label{eq:contact_direction_free}
\end{align}

\subsubsection{Safety and Virtual Constraints.}\label{sec:robot_safety}
The robot motion commands are projected to remain within the physical robot workspace and interpolated to ensure they are within the robot's dynamic operational space limits~\citep{Kroeger2010}. The force commands sent to the robot are similarly interpolated to ensure stability. The robot force command is saturated to maintain patient safety and comfort.
If the robot motion commands exceed the robot's reach, they are projected to remain within the physical robot workspace.

\subsection{Robotic Replay of Human Expert Scan}
\begin{figure*}[t]%
    \centering
    \includegraphics[width=\linewidth]{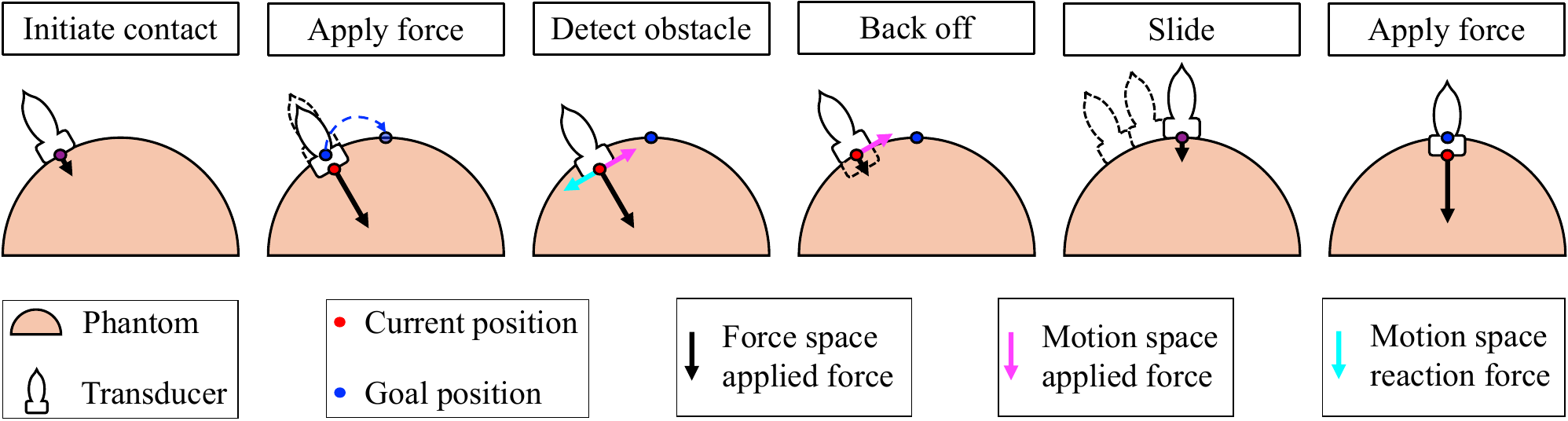}
    \caption{Force modulation strategy for applying axial force and generating tangential motion over the deformable phantom surface.}
    \label{fig:force_strategy}
\end{figure*}
When the robotic system is controlled to replay a freehand scan performed by a human expert,
the transducer motion generated and transducer force applied to acquire the resulting sequence of ultrasound images are used to parameterize a characteristic expert strategy. This strategy is used to generate the robot controller inputs.

\subsubsection{Recording Human Expert Strategies.}
A mechanism for recording the ultrasound transducer position and orientation, along with the direction and magnitude of force applied to the phantom, is presented in~\Cref{fig:freehand_scan} along with the freehand scan recording setup.
The mechanism features an internal housing to which the ultrasound transducer is attached. The internal housing is mounted to a force/torque sensor, enabling measurement of the contact forces between the transducer and phantom. An external adapter is placed above the force/torque sensor and motion capture markers are placed on the adapter. The radiologist holds
the external housing, which is fixed to the external adapter; therefore, the radiologist's gripping forces are not transmitted through the sensing-side of the force/torque sensor. The ultrasound images and radiologist's scan narration are also recorded.


Over the course of a detailed freehand ultrasound scan, a human expert generates redundant images by sweeping over the same areas of interest with a similar transducer orientation and applied force.
Two ways to reduce a detailed scan to a reference for a survey scan are by (1) removing scan sections with images generated earlier in the scan and (2) eliminating the transitions between different areas of interest.


The directions along which a human expert controls transducer motion and applied force are fundamental for understanding the expert's scanning strategies. To estimate the direction along which force is controlled, we construct a snapshot matrix of measured contact forces to extract the proper orthogonal decomposition (POD) basis~\citep{Kunisch1999}, with the assumption that the expert controls force mainly along one direction at any given time during an abdominal ultrasound scan. Specifically, at a given discrete time step $k \in \mathbb{Z}_{> 0}$, the current measured contact force, $f_k \in \mathbb{R}^3$, is combined with previous recorded contact forces to form the columns of the snapshot matrix, $S$. The amount of columns in the snapshot matrix is limited to a fixed positive integer, $n \in \mathbb{Z}_{> 0}$.
In summary,
\begin{align}
    S &=
    \begin{dcases}
        \begin{bmatrix}
            f_{k+1-n} & \cdots & f_{k}
        \end{bmatrix} \in \mathbb{R}^{3 \times n}
        & \textrm{if } k \geq n;
        \\
        \begin{bmatrix}
            f_{1} & \cdots & f_{k}
        \end{bmatrix} \in \mathbb{R}^{3 \times k}
        & \textrm{otherwise},
    \end{dcases} \label{eq:snapshot}
\end{align}
where $f_1$ is the first measured contact force.

SVD is performed on the contact force snapshot matrix to compute the decomposition $S = U \Sigma V^T$,
where $U$ encodes the directions corresponding to the singular values in the diagonal matrix $\Sigma$. The singular values
are listed in decreasing order ($\sigma_1 \geq \sigma_2 \geq \sigma_3$)
and normalized by a scale factor of $1/\sqrt{m}$, where $m$ is the number of columns in $S$, to make the force magnitude invariant of the number of force samples. The long contact force direction, $\hat{n}$, is computed as the column in $U$ corresponding to $\sigma_1$, the largest singular value in $\Sigma$.

\subsubsection{Robot Controller Implementation for Replay.}
When the robotic system replays a freehand scan by referencing the recorded expert strategy, the local robot controller follows the same structure as that presented in Section~\ref{sec:robot_controller}. Now,
\begin{itemize}
    \item The value of $\hat{n}$ used in~\eqref{eq:force_space_contact} to find the task specification matrices, $\Omega_f$ and $\Omega_m$, is estimated from SVD of the measured contact force snapshot matrix given by~\eqref{eq:snapshot};
    \item The expert's transducer position and orientation are used
    as the desired end-effector position and orientation, $x_{d,t}$ and $x_{d,r}$, to compute the control force in motion space given by~\eqref{eq:robot_controller_motion};
    \item The desired end-effector force is given directly by the expert's applied force, $F_a$, such that $F_d = F_a$ to find the control force in force space in~\eqref{eq:robot_controller_force}.
\end{itemize}

\subsubsection{Human-Inspired Force Strategy for Handling Elastic Surface Deformation.}
The specific phantom used for the experiments is made of a polyurethane elastomer that deforms significantly under applied pressure. Although the application of ultrasound gel reduces friction between the transducer face and phantom surface, with enough pressure, the surface deforms around the transducer and acts as an obstacle to tangential sliding motion.

From the recorded freehand ultrasound motion, we observed that the human expert applied more force with the transducer held relatively still, and applied less force while sliding the transducer to a new location on the surface. This strategy is incorporated into the robotic ultrasound system by modulating the end-effector's desired force as a function of its position error. Specifically, after finding the desired end-effector force, $F_d$, this force is blended between its nominal value and a smaller positive threshold, $F_{d,min}$, as follows:
\begin{align}
    F_d &\coloneqq \alpha F_{d,min} + (1-\alpha) F_d,
\end{align}
where
\begin{align}
    \alpha &= 1 - \exp \Bigl( -\beta \bigl|\bigl|\Omega_{m,t}
    e_{m,t}
    \bigr|\bigr| \Bigr).
\end{align}
This exponential smoothing with decay rate $\beta$ allows the robot to reduce its applied force when diverging from its desired position. In turn, the reduced force results in less local surface deformation and facilitates tangential sliding over the surface. The positive force bias $F_{d,min}$ helps the end-effector maintain constant contact with the surface.

The force strategy for handling elastic surface deformation is illustrated in~\Cref{fig:force_strategy}. After the robot applies a force to the phantom at a static location, the surface deforms around the transducer and creates an obstacle to tangential motion. When the desired location is updated, the robot detects said obstacle and reduces its applied force until the end-effector slides to the new desired location. The nominal desired force is applied once the end-effector reaches the goal position.

\subsubsection{Comparing Freehand and Robotically-acquired Images.}\label{sec:image_comparison}
Distance-based hashing is a method used to compare a given pair of images~\citep{Athitsos2008}. This approach reduces the size of the input image to $9 \times 8$ pixels to remove high frequencies and detail, converts the result to a grayscale picture, computes the relative gradient direction by checking the difference between adjacent pixels, and assigns bits to the resulting $64$ pixel differences. The computed set of bits is the image's distance-based hash value, which represents the relative change in image brightness intensity.

In ultrasound imaging contexts, the Hamming distance, or the number of different corresponding bits, between two image hashes has been used to check for image similarity~\citep{Senthilkumar2016}. Dissimilar images yield large Hamming distances, while a Hamming distance of $0$ implies the two images are identical.

%% file: sections/4-enabling_autonomy.tex
\section{Enabling Autonomous Robotic Ultrasound with Vision and Touch}\label{sec:autonomy}
To advance towards autonomous scanning capabilities, the robotic ultrasound imaging platform must be able to independently generate and execute scanning paths.
These paths are designed to perform a comprehensive survey scan of the abdominal phantom by scanning the anatomies located underneath the rib cage as well as those directly beneath the soft tissue. The paths also ensure that the transducer navigates around anatomical constraints, such as the curvature of the ribs and soft tissue, while maintaining constant contact with the surface.

\begin{figure}[t]%
    \centering
    \includegraphics[width=\linewidth]{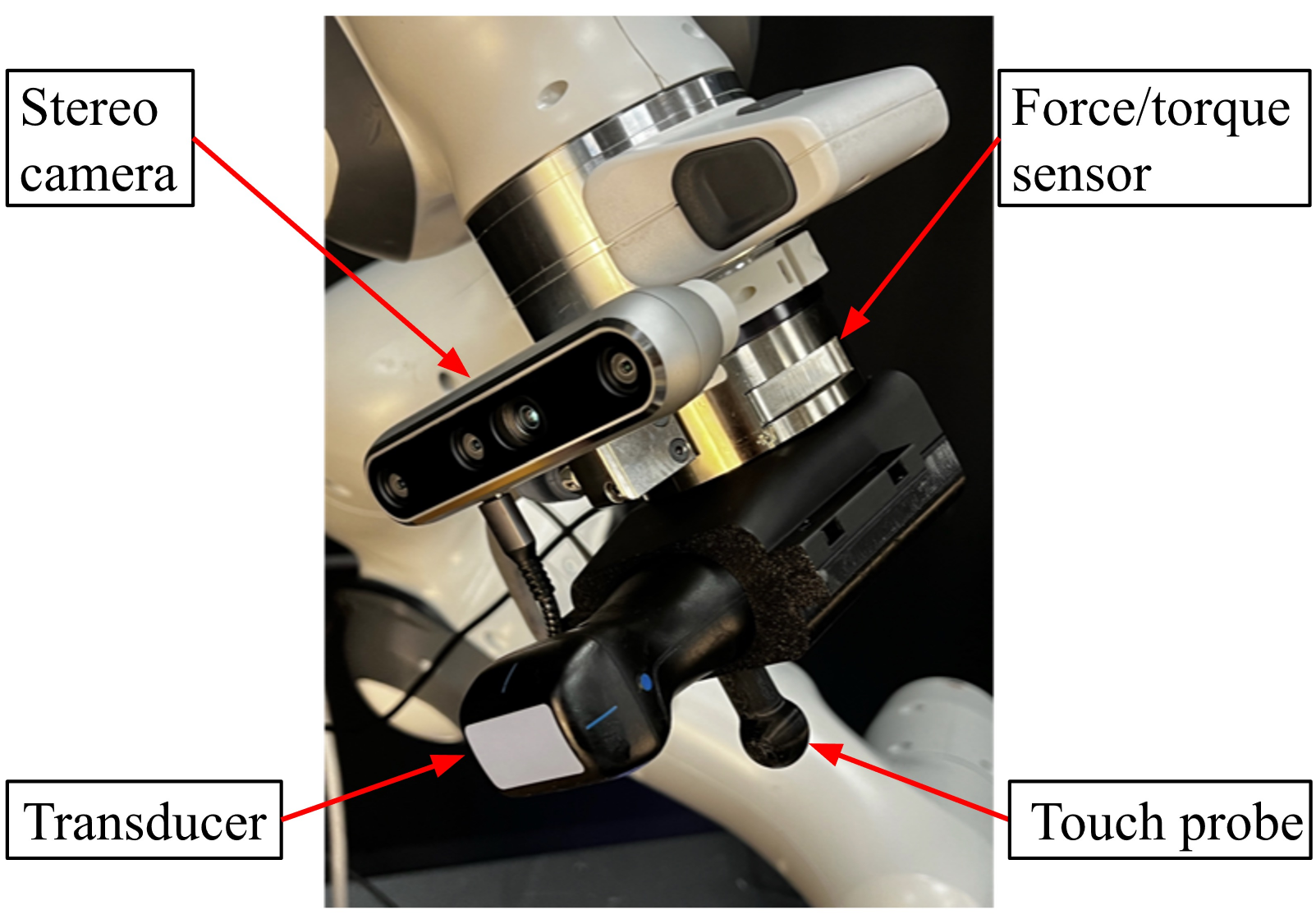}
    \caption{End-effector design for autonomous robotic ultrasound imaging. The stereo camera helps compute initial scanning paths that are specific to the patient's topography. The touch probe is used to refine the scanning paths by delineating the boundary between the ribs and soft tissue.}
    \label{fig:auto_ee}
\end{figure}

A novel end-effector design for autonomous robotic ultrasound scan acquisition is presented in~\Cref{fig:auto_ee}. A stereo camera is incorporated into the robotic system to provide estimates of the phantom's position and surface topography, which are used to plan the initial scanning paths. Then, the paths are refined by using the touch probe and force/torque sensor to measure the phantom stiffness at key locations and delineate the boundary between the rib cage and the abdominal soft tissue.
The force/torque sensor is also used to maintain constant contact between the ultrasound transducer and phantom surface.

\subsection{Vision for Patient-Specific Ultrasound Path Planning}\label{sec:vision}

In order to generate the initial patient-specific scan paths, an external vision system is incorporated to capture stereo images of the medical phantom. The stereo images provide a point cloud encoding the estimated position and color of each point in three-dimensional space. The input point cloud is processed to generate a watertight mesh representing an estimate of the physical phantom surface.

\begin{figure*}[t]%
    \centering
    \includegraphics[width=\linewidth]{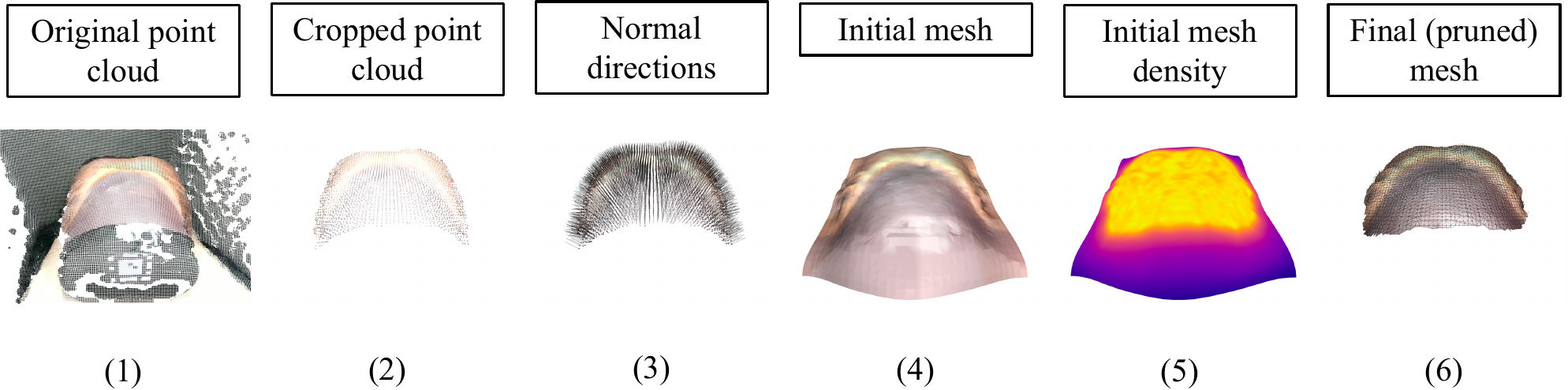}
    \caption{Generating a watertight surface mesh representing the phantom surface from an input point cloud. \textbf{Left to right}: (1) The input point cloud contains the phantom; (2) the point cloud is cropped to the phantom boundaries; (3) the normal direction is computed at each point; (4) a watertight mesh is generated with Poisson surface reconstruction; (5) the density map represents the mesh regions supported by the cropped point cloud; (6) low-density regions in the initial mesh are pruned to get the output mesh.}
    \label{fig:pcd_to_mesh}
\end{figure*}

The procedure for generating the watertight surface mesh is shown in~\Cref{fig:pcd_to_mesh}.
First, the original point cloud is cropped to the physical extents of the abdominal phantom. The point cloud is then downsampled and the normal direction at each point in the cloud is estimated by finding the normal vector to the plane that best fits the point's $k$-nearest neighbors. All resulting point cloud normal vectors are oriented towards the camera viewpoint to resolve ambiguity in the normal direction. Poisson surface reconstruction is applied to generate a watertight surface mesh from the cropped point cloud and estimated normal directions~\citep{Kazhdan2006}. Each cell in the octree used for Poisson surface reconstruction records the number of input points contributing to it, creating a density estimate for each region in the reconstructed surface. This density map is used to prune low-density regions, removing extrapolated parts of the mesh that are not supported by the point cloud. The output surface mesh is generated by removing duplicate vertices and non-manifold edges from the pruned mesh.

\begin{figure}[t]%
    \centering
    \includegraphics[width=\linewidth]{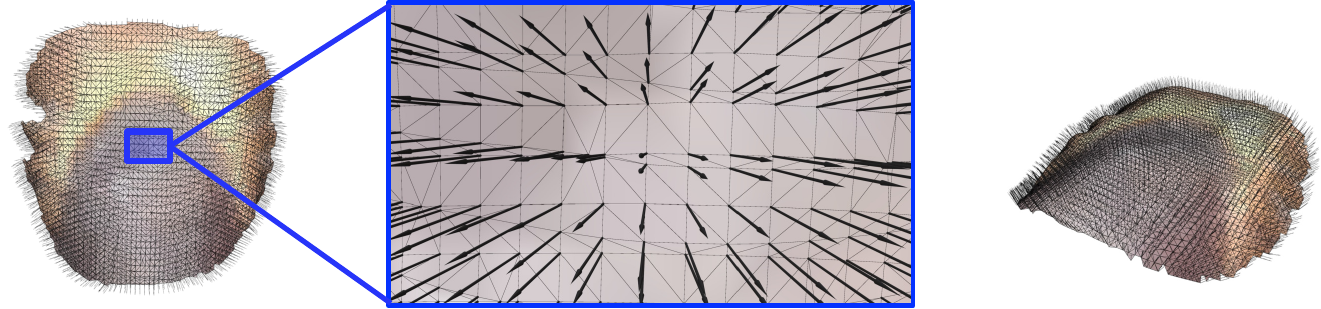}
    \caption{Watertight surface mesh representing the abdominal phantom surface. The closeup image highlights the vertex normal vectors. This mesh provides a continuous representation of the position and normal direction at any point on the surface.}
    \label{fig:mesh_normals}
\end{figure}

The final mesh provides a continuous representation of the position and orientation at every location on the phantom surface. Specifically, for any point on a given mesh face, the point's position and normal direction are interpolated from the face vertices by using the point's barycentric coordinates. The mesh vertices, edges, vertex normals, and face colors are illustrated in~\Cref{fig:mesh_normals}.

Given an ultrasound transducer with a specified geometry and a mesh representing the abdominal phantom surface, initial ultrasound scan paths are computed to sweep over the surface with full coverage. Concretely, the transducer face is treated as a planar rectangle. A serpentine path that sweeps the rectangular transducer face over the surface mesh is computed by Algorithm~\ref{algo:plan_path}.

\begin{algorithm}[t]
\caption{Serpentine Path Generation on Surface Mesh}
\label{algo:plan_path}
\begin{algorithmic}[1]\raggedright

\Require Mesh $M$, sweep width $w$, longitudinal axis $\hat{u}_{long}$, lateral axis $\hat{u}_{lat}$
\Ensure Set of geodesic paths $P$ sweeping over mesh



\State $v_{long,min} \gets$ vertex with minimal coordinate in $M$ along $\hat{u}_{long}$
\State $v_{long,max} \gets$ vertex with maximal coordinate in $M$ along $\hat{u}_{long}$
\State $P \gets \emptyset$ \Comment{Initialize empty path set}
\State $v_{long} \gets v_{long,max}$ \Comment{Initialize longitudinal vertex}
\State $\hat{d}_{lat}$ $\gets$ $+\hat{u}_{lat}$ \Comment{Initialize lateral sweep direction}

\While{$(v_{long} - w) > v_{long,min}$}
    \State $R \gets$ set of vertices $v \in M$ where \\
    $\bigl( v_{long} - w \bigr) \leq \bigl( v \cdot \hat{u}_{long} \bigr) < v_{long}$
    \State $v_{lat,min} \gets$ vertex with minimal coordinate in $R$ along $\hat{d}_{lat}$
    \State $v_{lat,max} \gets$ vertex with maximal coordinate in $R$ along $\hat{d}_{lat}$
    


    \If{$P$ is non-empty}
        \State $\pi_{long} \gets$ \Call{computePath}{$v_{lat,prev}, v_{lat,min}$}
        \State $P \gets$ \Call{appendPath}{$P, \pi_{long}$}
    \EndIf
    \State $\pi_{lat} \gets$ \Call{computePath}{$v_{lat,min}, v_{lat,max}$}
    \State $P \gets$ \Call{appendPath}{$P, \pi_{lat}$}
    \State $v_{lat,prev} \gets v_{lat,max}$ \Comment{Store for next connection}
    \State $v_{long} \gets v_{long} - w$ \Comment{Move to next lateral region}
    \State $\hat{d}_{lat}$ $\gets$ $-\hat{d}_{lat}$ \Comment{Flip lateral sweep direction}
\EndWhile

\State \Return $P$
\end{algorithmic}
\end{algorithm}

\begin{figure}[t]%
    \centering
    \includegraphics[width=\linewidth]{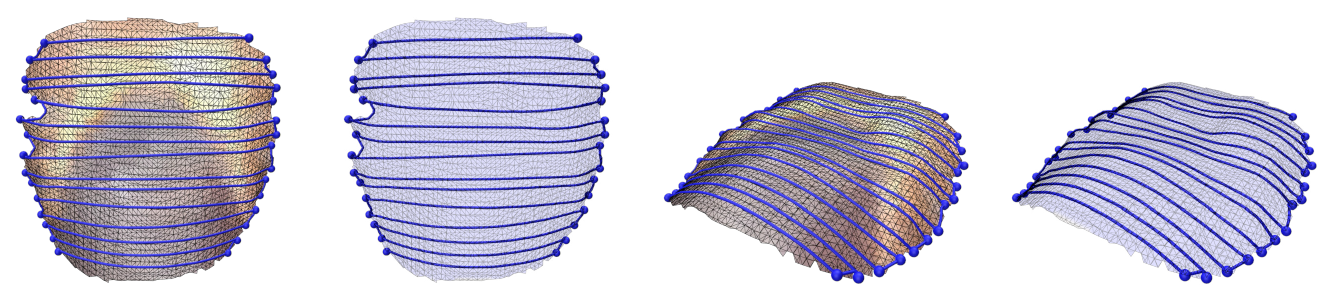}
    \caption{Construction of the sagittal ultrasound sweep. The computed sweep paths alternate from right-to-left and left-to-right while moving from head-to-toe. Each path is the geodesic connecting the start and end path vertices.}
    \label{fig:mesh_dexter_path}
\end{figure}

The initial ultrasound sweep path in the sagittal direction is found by applying Algorithm~\ref{algo:plan_path} with the longitudinal and lateral sweep axes set to the phantom's head-to-toe and right-to-left directions, respectively. Note that the function \textproc{computePath} finds the geodesic path on the surface between the start and end vertices using the technique described by~\cite{Sharp2020}.
The resulting sweep path is demonstrated in~\Cref{fig:mesh_dexter_path}.

\subsection{Refining Vision-based Ultrasound Paths with Touch}

Although the initial sagittal sweep provides full coverage of the transducer face over the abdominal phantom, there are sections of the sweep that pass directly over the rib cage. This is undesirable due to acoustic shadowing, where ultrasound waves are blocked by the dense and highly reflective bone of the ribs, creating dark, non-diagnostic areas beneath them.

\begin{figure*}[t]%
    \centering
    \includegraphics[width=\linewidth]{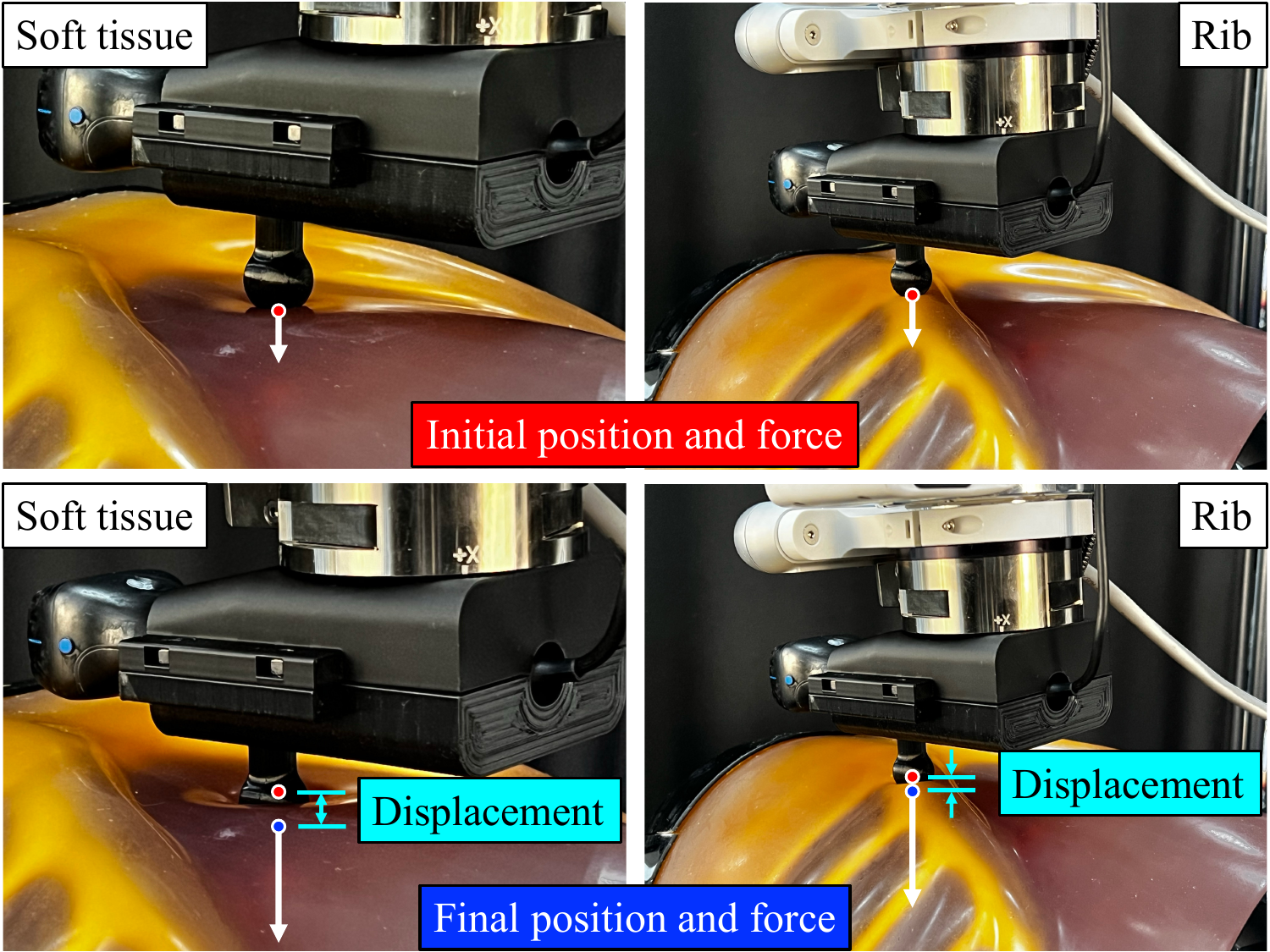}
    \caption{Procedure for measuring linear stiffness at different points on the phantom. At each point, a known force is applied and the resulting linear displacement is measured. For the same applied force, rib points displace less than soft tissue points.}
    \label{fig:rib_detection}
\end{figure*}

The boundary between the rib cage and soft abdominal tissue must be delineated to refine the vision-based ultrasound scanning paths such that they do not pass over the rib cage. One method for delineating the rib cage boundary is by physically probing the phantom surface since the bones of the rib cage are significantly stiffer than surrounding soft tissues. Specifically, a known applied force is applied to different regions on the phantom and the resulting displacement is measured. The linear stiffness is computed from the force and displacement measurements. The rib regions are classified by stiffness measurements above a specific threshold, and soft tissue regions are represented by stiffness measurements below said threshold.
The experimental setup for measuring linear stiffness at different points on the phantom is illustrated in~\Cref{fig:rib_detection}.

\subsubsection{Endowing Ultrasound Paths with Human Expert Strategies.}

One strategy that radiologists employ to aim the ultrasound waves away from the ribs and access the underlying anatomies is to angle the transducer underneath the rib cage.
This technique improves imaging quality by avoiding shadowing artifacts and ensuring the ultrasound waves reach the target anatomy unobstructed. This approach is particularly important for imaging structures like the liver, heart, and lungs, which are located beneath the rib cage.

In the soft tissue region, radiologists typically angle the ultrasound transducer normal to the patient's surface to maximize ultrasound wave penetration and signal strength. When the transducer is perpendicular, the ultrasound waves
reflect back to the transducer more efficiently, producing a stronger echo and clearer image. This orientation minimizes artifacts like refraction and improves visualization of underlying structures. Additionally, a normal angle helps achieve uniform contact with the patient's skin, reducing the risk of air gaps that disrupt ultrasound wave transmission.

\begin{figure*}[t]%
    \centering
    \includegraphics[width=\linewidth]{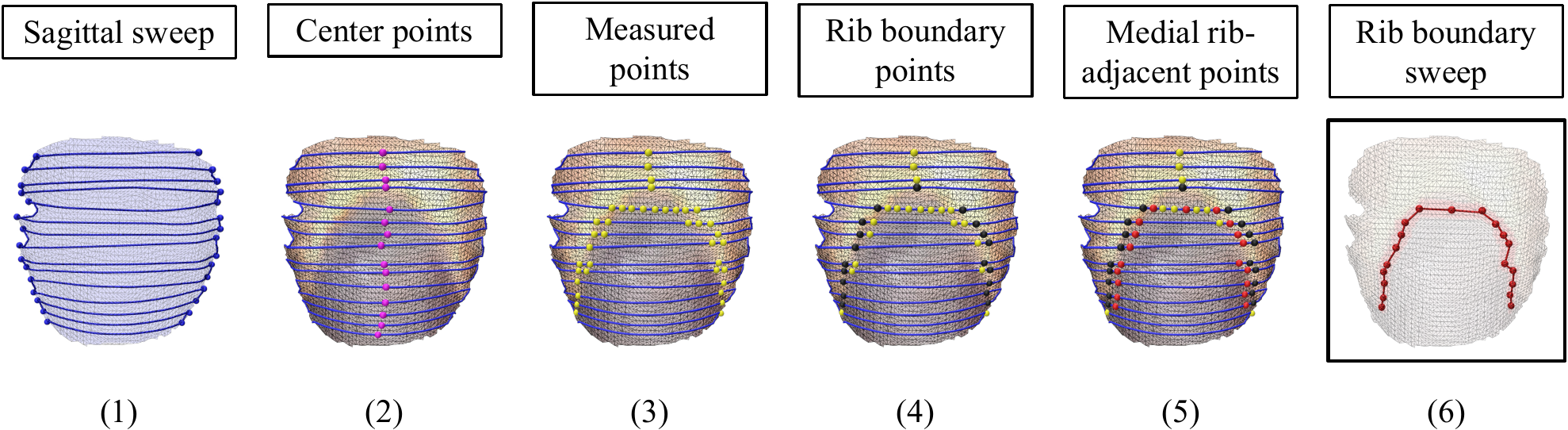}
    \caption{Delineating the rib cage boundary and generating the under-rib sweep. \textbf{Left to right}: (1) The sagittal sweep path provides the reference for the rib detection procedure; (2) the center of each lateral path is found; (3) the stiffness is measured at key points along each lateral path; (4) the measured points corresponding to rib are identified (black); (5) the points adjacent to the rib points are found (red), representing the rib boundary; (6) the under-rib path is composed of geodesics connecting the rib boundary points.}
    \label{fig:mesh_rib_sweep}
\end{figure*}

We use the delineation of the rib cage boundary to enable these two human expert strategies. In particular, we start with the sagittal sweep presented in Section~\ref{sec:vision}.
Then, we follow the procedure described in Algorithm~\ref{algo:delineate_ribs} to identify the vertices on the mesh corresponding to the rib cage boundary.
The first measured point is at the top and center of the sternum, which provides the initial reference for the rib stiffness. The stiffness at the center of the following lateral paths in the sagittal path are measured and compared to a soft-tissue--to--rib stiffness ratio, which we set to $0.8$. If the measured stiffness is above the rib stiffness threshold, the current measurement is incorporated in the average rib stiffness and the center point on the next lateral path is measured. This process repeats until soft tissue is detected. Then, the stiffness measured at points to the left and right of the soft tissue center point are compared to the soft tissue stiffness determined by the first soft tissue point. The points to the left and right of each center point in the sagittal path are spaced apart by a fixed distance on each lateral path, which we set to $\SI{10}{\milli\meter}$. The procedure leverages prior knowledge of the rib cage geometry such that, whenever a rib boundary point is detected, the procedure does not restart at the center of the next lateral path in the sagittal sweep. Instead, the search for the next rib boundary point starts underneath the previously detected rib boundary point. This search continues on both sides of the rib cage until the lateral abdominal boundaries are reached.
To enable the sweep angled underneath the rib cage, we construct a geodesic path between the rib boundary vertices. Each step of this process is shown in~\Cref{fig:mesh_rib_sweep}.

\begin{algorithm*}[t]
\caption{Rib Boundary Delineation and Path Generation}
\label{algo:delineate_ribs}
\begin{multicols}{2}
\begin{algorithmic}[1]\raggedright

\Require Mesh $M$,
soft-tissue--to--rib stiffness ratio $\eta$,
sagittal sweep lateral paths $\pi_1 \dots \pi_N$
\Ensure Set of geodesic paths $P$ through rib cage boundary

\State $c_1 \gets$ \Call{centerPoint}{$\pi_1$} \Comment{Center point at sternum}
\State $k \gets$ \Call{measureStiffness}{$c_1$}
\State $k_{rib} \gets k$ \Comment{Use sternum as initial rib stiffness}
\State $k_{prev} \gets 0$ \Comment{Store for computing average stiffness}
\State $i \gets 1$ \Comment{Start at first lateral path}

\LComment{Iterate over center points until soft tissue is detected}
\While{$k \geq \bigl( \eta \cdot k_{rib} \bigr)$}
\State $k_{rib} \gets \frac{k_{prev} + k}{i}$ \Comment{Compute average rib stiffness}
\State $k_{prev} \gets k$
\State $c_i \gets$ \Call{centerPoint}{$\pi_i$}
\State $k \gets$ \Call{measureStiffness}{$c_i$}
\State $i \gets i + 1$ \Comment{Move to next lateral path}
\EndWhile

\State $k_{soft} \gets k$ \Comment{First non-rib stiffness is soft tissue}
\State $k_l, k_r \gets k_{soft}$
\State $b_{l}, b_{r} \gets c_{i-1}$ \Comment{Initialize left/right boundary point}
\State $s_l, s_r \gets \SI{0}{\milli\meter}$ \Comment{Track left/right shift from center point}

\LComment{Identify rib boundary on the remaining lateral paths}
\While{$i \leq N$}

    \State $c_i \gets$ \Call{centerPoint}{$\pi_i$}
    \State $b_{l,prev} \gets b_{l}$ \Comment{Store for next connection}
    \State $b_{r,prev} \gets b_{r}$

    \LComment{Measure left-of-center points until rib detected}
    \While{$k_l < k_{soft}$}
    \State $k_l \gets$ \Call{measureStiffness}{$c_i - s_l$}
    \State $s_l \gets s_l + \SI{10}{\milli\meter}$ \Comment{Shift away from center}
    \EndWhile
    
    \LComment{Measure right-of-center points until rib detected}
    \While{$k_r < k_{soft}$}
        \State $k_r \gets$ \Call{measureStiffness}{$c_i + s_r$}
        \State $s_r \gets s_r + \SI{10}{\milli\meter}$ \Comment{Shift away from center}
    \EndWhile

    \LComment{Set medial rib-adjacent points as rib boundary}
    \State $b_{l} \gets c_i - (s_l - \SI{10}{\milli\meter})$
    \State $b_{r} \gets c_i + (s_r - \SI{10}{\milli\meter})$

    \State $\pi_l \gets$ \Call{computePath}{$b_{l,prev}, b_{l}$}
    \State $\pi_r \gets$ \Call{computePath}{$b_{r,prev}, b_{r}$}
    \State $P \gets$ \Call{appendPath}{$P, \pi_{l}, \pi_{r}$}

    
    \State $i \gets i + 1$ \Comment{Move to next lateral path}
\EndWhile



\State \Return $P$
\end{algorithmic}
\end{multicols}
\end{algorithm*}

\begin{figure}[t]%
    \centering
    \includegraphics[width=\linewidth]{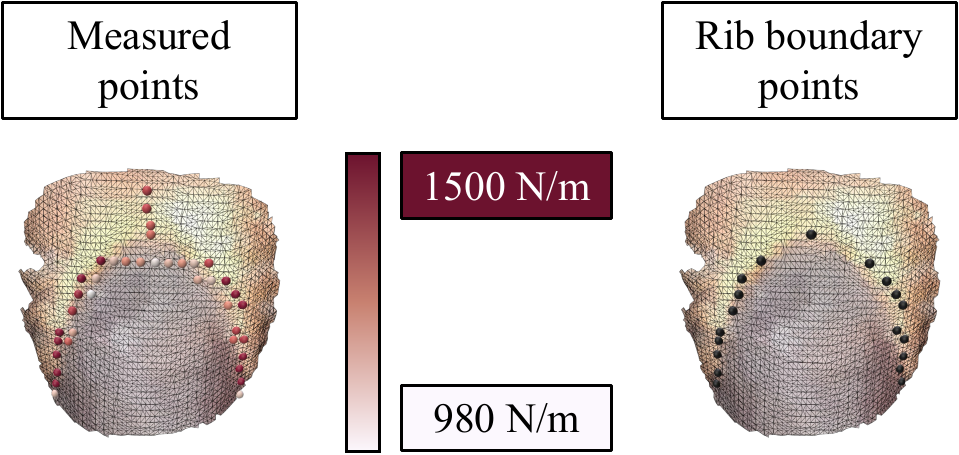}
    \caption{Linear stiffness values at the measured points on the phantom and the corresponding points on the rib cage boundary.
    }
    \label{fig:measured_stiffnesses}
\end{figure}
\begin{figure}[t]%
    \centering
    \includegraphics[width=\linewidth]{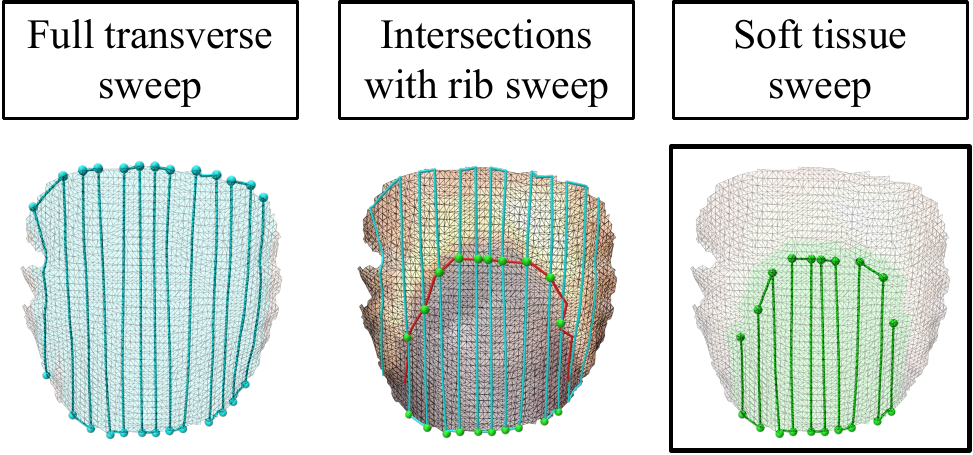}
    \caption{Generating the soft tissue transverse sweep
    by removing the portions of the full transverse sweep that are above the rib cage boundary.
    }
    \label{fig:mesh_transverse_sweep}
\end{figure}


The linear stiffness values at the points on the phantom measured in the rib boundary delineation procedure are shown in~\Cref{fig:measured_stiffnesses}. The points classified as those on the rib boundary are also illustrated.

After delineating the rib boundary, we find an initial transverse sweep covering the mesh, which is generated similarly to the sagittal sweep by changing the longitudinal and lateral sweep axes. We use the rib boundary to separate this transverse sweep into the portions that sweep over the (1) rib cage and (2) soft tissue. Specifically, the intersections between the transverse and rib boundary sweeps are found. Then, the portion of the transverse sweep above the rib boundary is removed, leaving only the portion sweeping over the soft tissue as shown in~\Cref{fig:mesh_transverse_sweep}.

\subsubsection{Compliant Frame Parameterization.}
With the unified motion and force controller in operational space described in~\Cref{sec:robot_controller}, the robot end-effector tracks a compliant frame parameterized by desired motion and force. In motion space, the end-effector tracks a desired position and orientation. Similarly, in force space, the end-effector tracks a desired force and moment. Since the end-effector rigidly grasps the transducer, the desired motion/force of the end-effector and transducer are used interchangeably.

The two final sweep paths provide coverage of both the under-rib and soft tissue abdominal areas. The desired transducer face position is parameterized as a function of time with a given desired linear transducer velocity.

The notation for specifying the desired transducer orientation is as follows. The transducer's normal direction is denoted $\hat{n}$ and points from the transducer face to the tail end of the transducer. By convention, the transducer's binormal direction, $\hat{b}$, is always along the smaller axis of the transducer face plane. The tangent direction is derived from the cross product between the binormal and normal directions, $\hat{t} = \hat{b} \times \hat{n}$.

\begin{figure*}[t]%
    \centering
    \includegraphics[width=\linewidth]{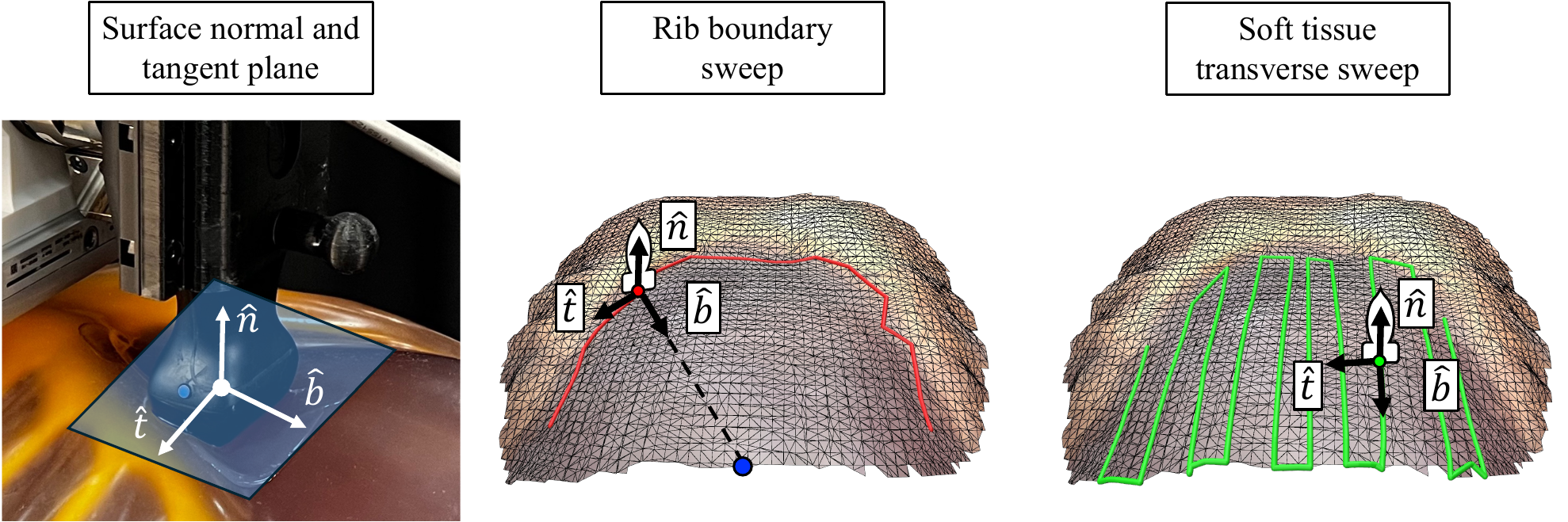}
    \caption{Specifying desired transducer orientation for autonomous ultrasound sweeps. \textbf{Left}: The transducer's normal, binormal, and tangent directions all intersect at the center of the transducer face.
    \textbf{Middle}: To view under the ribs, the binormal direction points from the transducer to the bottom-most lateral center point on the mesh (blue). The desired transducer normal direction is rotated from the surface normal to view under the ribs.
    \textbf{Right}: Along the soft tissue, the binormal direction points from head-to-toe to acquire images in the transverse plane. The desired transducer normal direction is simply the surface normal.}
    \label{fig:auto_ori}
\end{figure*}

The procedure for specifying the desired transducer orientation differs between the under-rib and soft tissue sweep paths. Both procedures are illustrated in~\Cref{fig:auto_ori}.

To angle the transducer under the ribs and parallel to the rib boundary during the under-rib sweep, the nominal binormal direction points towards the bottom-most lateral center point on the mesh. The desired normal direction is given by rotating the surface normal at the transducer's desired position by $\SI{-30}{\deg}$ about $\hat{t}$. Then, $\hat{b}$ is computed by projecting the nominal binormal direction onto the plane perpendicular to $\hat{n}$.

In contrast, the desired value of $\hat{n}$ for the soft tissue sweep is simply the surface normal at the transducer's desired position. To acquire images in the transverse plane, the unprojected binormal direction is always set to point from head-to-toe on the phantom. Then, this vector is projected onto the plane perpendicular to $\hat{n}$ to find $\hat{b}$.

For simplicity, the desired contact force between the transducer face and phantom surface is set to a constant value of $\SI{15}{\N}$, which was approximately the maximum force applied by the radiologist during the freehand scan. The end-effector orientation is controlled entirely in the motion space; therefore, there is no desired moment commanded to the robot.

%% file: sections/5-experimental_results.tex
\section{Experimental Results}\label{sec:results}
Three different scans of the abdominal ultrasound phantom
were performed: (1) a freehand scan performed by an expert radiologist, (2) a robotic replay scan based on the freehand scan data, and (3) an autonomous robotic scan. The autonomous robotic scan was split into two sweeps: one looking underneath the rib cage boundary, and the other viewing normal to the soft abdominal tissue.

All scans were performed using an ultrasound phantom containing a liver, gallbladder, pancreas, kidneys, and detailed vascular structures. The phantom also contained lesions including cysts, hypoechoic lesions, and tumors. The motion and force data presented in this section is with respect to the phantom's RAS anatomical coordinate frame, which is the one illustrated in~\Cref{fig:system}.

\subsection{Freehand Baseline Scan}
The radiologist performed a freehand detailed ultrasound scan of the entire abdomen using the mechanism for recording the ultrasound transducer motion and force applied to the phantom described earlier in this paper.

The full freehand ultrasound scan trajectory was reduced to avoid redundant images and remove transitions between the key areas of interest identified in the radiologist's audio narration recorded during the scan. The full and reduced scan trajectories are presented on the right of~\Cref{fig:freehand_scan}.
The reduction of the full scan is measured with respect to the total scan time and distance covered by the transducer. The full scan was performed in $\SI{362}{\s}$ and the reduced scan time is $\SI{172}{\s}$, which is a $\SI{52.5}{\percent}$ reduction in the total scan time.
The total scan distance covered was $\SI{27.3}{\meter}$ in the freehand scan and $\SI{3.03}{\meter}$ in the reduced scan, which is a $\SI{88.9}{\percent}$ reduction.

\begin{figure*}[t]%
    \centering
    \includegraphics[width=\linewidth]{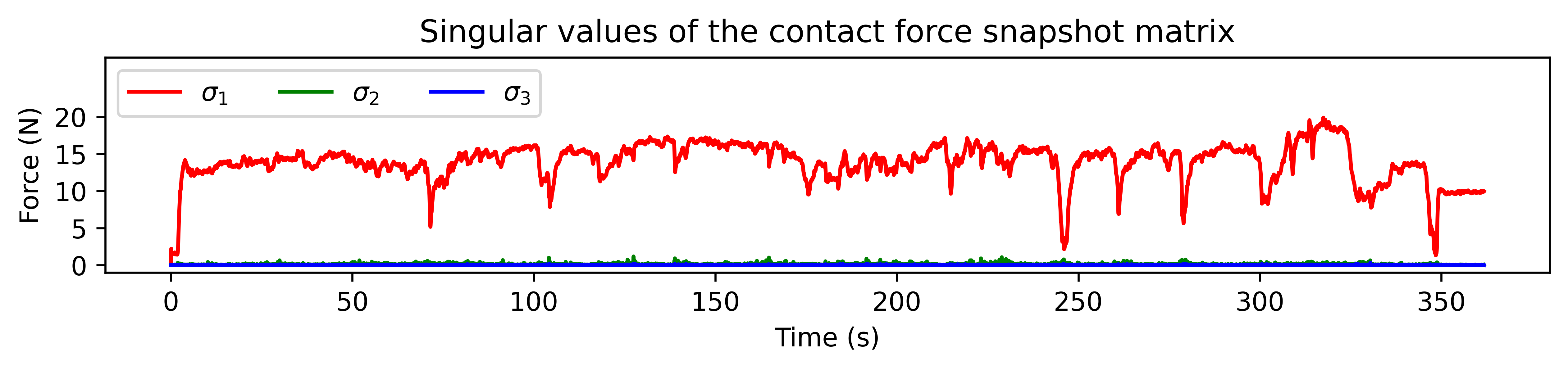}
    \caption{Singular values of the contact force snapshot matrix from the full freehand scan, showing that the expert controls force along a single direction.}
    \label{fig:singular_values}
\end{figure*}

An assumption made when extracting the contact force direction using POD is that the radiologist applies force along a single direction. This assumption is tested by investigating the three singular values computed at each time step in the freehand scan, which are shown in~\Cref{fig:singular_values}.
The average values of the three singular values were, in descending order, $\SI{13.90}{\N}$, $\SI{0.12}{\N}$, and $\SI{0.03}{\N}$. The average ratio between the largest and second-largest singular values was $\SI{217}{\percent}$, while the average ratio between the largest and smallest singular values was $\SI{721}{\percent}$.

\subsection{Robotic Replay of Freehand Scan}
A replay scan was performed by the robot, in which the transducer motion and force recorded during the baseline freehand scan were used as the basis for inputs to the robot controller. The inputs were optimized to remove redundant scan segments, particularly in portions where the radiologist slowed down to explain specific pathologies. The transducer motion, contact force between the transducer and phantom, and generated ultrasound images were recorded.

\subsubsection{Motion and Force Tracking Performance.}
\begin{figure*}[t]%
    \centering
    \includegraphics[width=\linewidth]{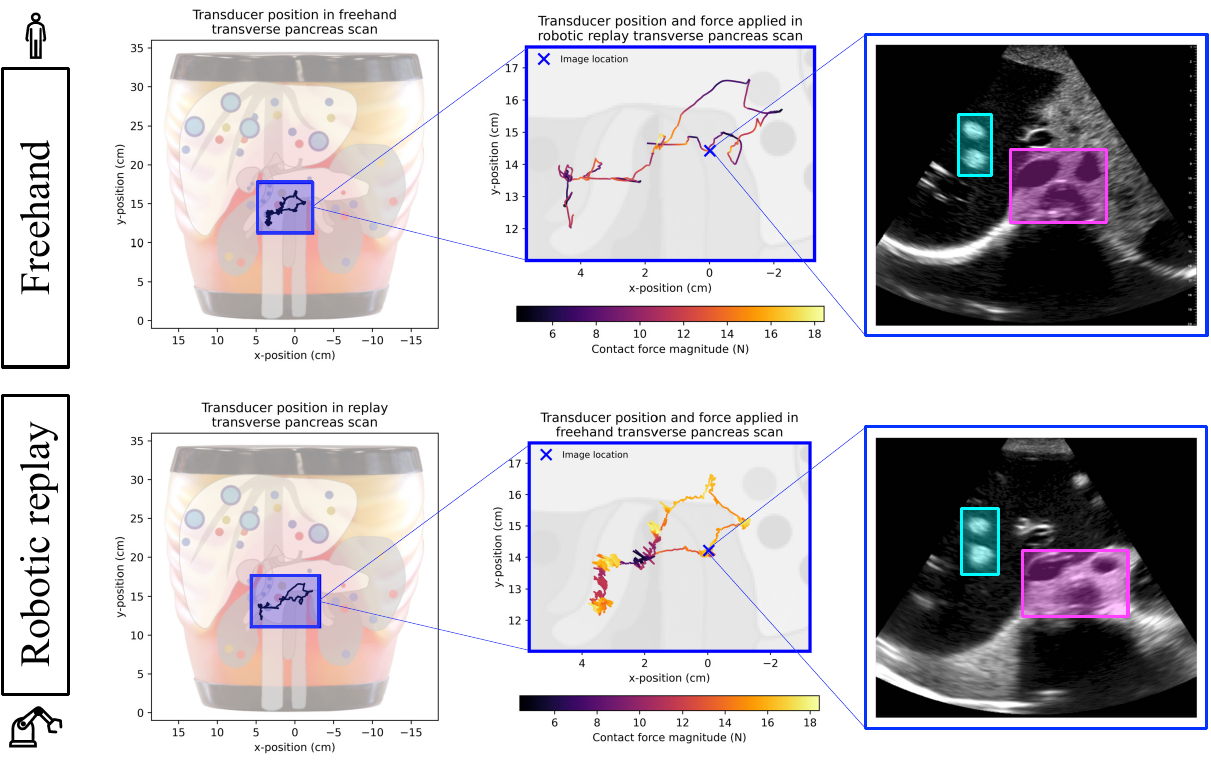}
    \caption{Ultrasound scans of the pancreas in the transverse plane.
    \textbf{Left}: The transducer motion.
    \textbf{Middle}: The transducer motion with
    the contact force magnitude applied onto the phantom at each point.
    \textbf{Right}: The ultrasound image acquired at the blue X location. Two echogenic lesions are highlighted in cyan, while hypoechoic lesions are highlighted in magenta.}
    \label{fig:functional_equivalence}
\end{figure*}

The following results pertain to the portion of the freehand and robotic replay scans focused on imaging the pancreas in the transverse plane.
The transducer motion and force
during this portion of the freehand and robotic replay scans
are shown in~\Cref{fig:functional_equivalence},
along with the images acquired at the most similar locations between the two scans.
The same set of echogenic and hypoechoic lesions are visible in both images.

\begin{figure*}[t]%
    \centering
    \includegraphics[width=\linewidth]{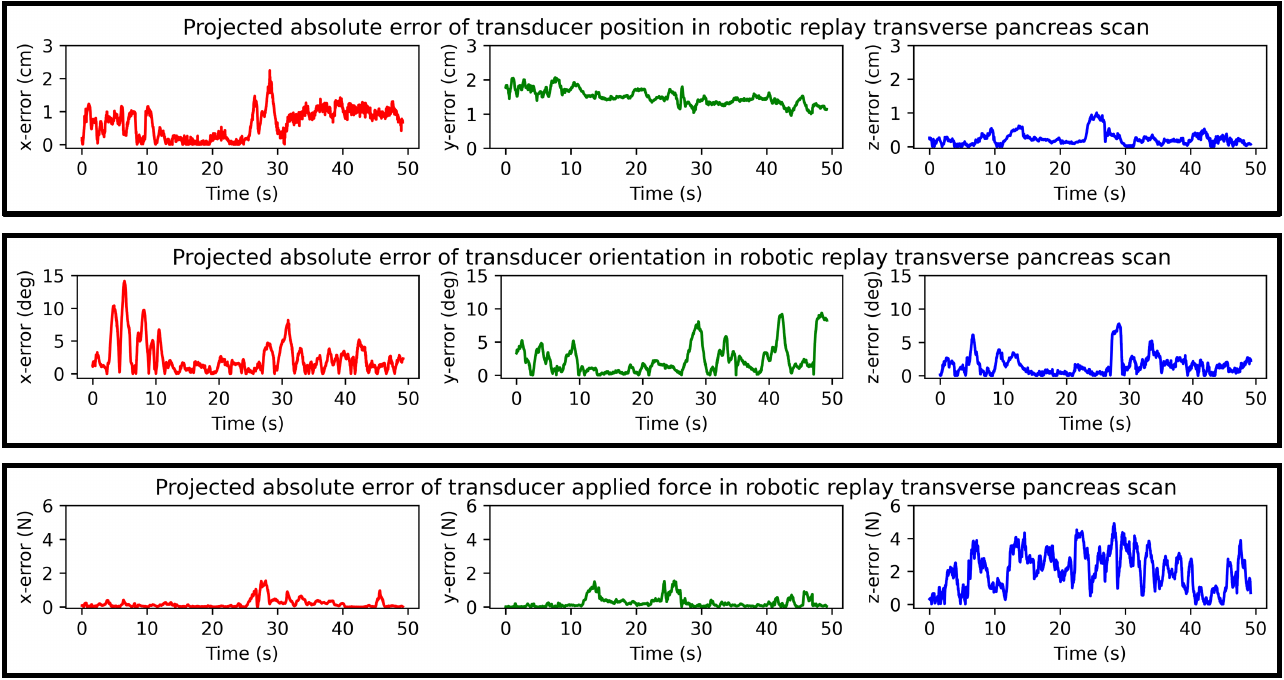}
    \caption{Projected transducer motion and force error during robotic replay scan.}
    \label{fig:replay_projected_errors}
\end{figure*}

The error magnitudes between the freehand and robotic linear and rotational motion, as well as linear applied force, projected into the extracted motion and force spaces are shown in~\Cref{fig:replay_projected_errors}.

The average motion space position error magnitude along the $x$, $y$, and $z$-directions were $\SI{0.70}{\cm}$, $\SI{0.44}{\cm}$, and $\SI{0.50}{\cm}$, respectively. The average orientation error magnitude about the $x$, $y$, and $z$-directions were $\SI{2.4}{\deg}$, $\SI{1.6}{\deg}$, and $\SI{2.3}{\deg}$, respectively. The average force space linear force error magnitude along the projected $x$, $y$, and $z$-directions were $\SI{0.20}{\N}$, $\SI{0.28}{\N}$, and $\SI{2.20}{\N}$, respectively.

\subsubsection{Image Comparisons.}
\begin{figure*}[t]%
    \centering
    \includegraphics[width=\linewidth]{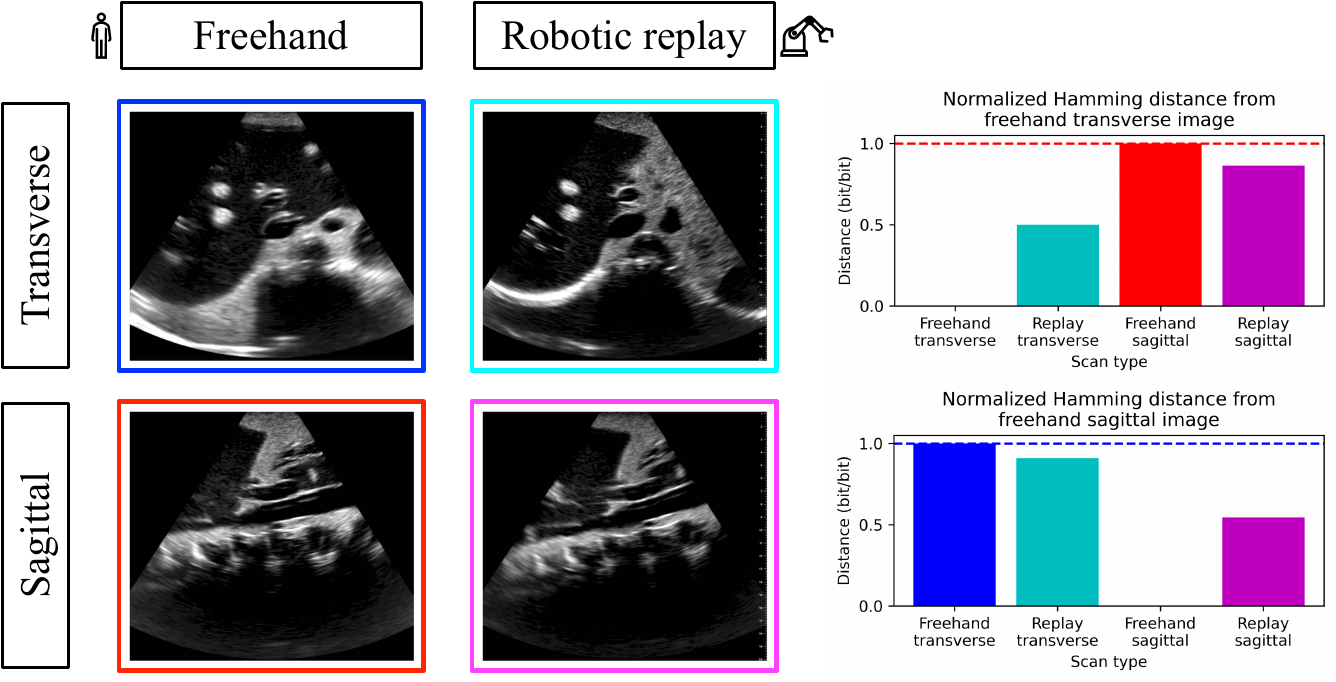}
    \caption{Comparing freehand and robotically-acquired ultrasound images.
    \textbf{Left}: The freehand and robotic replay scan images in the transverse and sagittal planes.
    \textbf{Right}: The Hamming distance between distance-based image hashes, normalized by the distance between the two different freehand images.}
    \label{fig:dhash}
\end{figure*}

The following results focus on ultrasound images of the pancreas in the transverse plane and right lobe of liver in the sagittal plane taken at the most similar locations in the freehand and robotic replay scans.
Image similarity is measured using the Hamming distance between distance-based image hashes, as described in Section~\ref{sec:image_comparison}.
The Hamming distance between the two different freehand images is used as the normalization factor. The different images along with the normalized Hamming distance of all four images with respect to each freehand image are shown in~\Cref{fig:dhash}.
The normalized Hamming distance between the two different freehand images is exactly $1$, whereas the distance is $0$ between the same two freehand images.
The normalized Hamming distances from the freehand transverse and sagittal images to the corresponding robotically-acquired images are $\SI{72.7}{\percent}$ and $\SI{66.7}{\percent}$ less than the distances from the freehand images to the opposite robotic images.

Importantly, for a given anatomy, the same features are identified in both the freehand and robotically-acquired scan images. The same set of lesions are visible in both scan images of the pancreas in the transverse plane. The inferior vena cava and posterior vein are both visible in the scan images of the right lobe of liver in the sagittal plane.


\subsection{Autonomous Robotic Scan}
The robot performed an autonomous scan composed of two different sweeps: one under the rib cage boundary, and the other directly over the soft abdominal tissue. The autonomous transducer motion, contact force between the transducer and phantom, and generated ultrasound images were recorded. 
The desired transducer velocity was set to $\SI{0.01}{\meter\per\second}$ for both autonomous sweeps. The current ultrasound image was tracked with the corresponding transducer pose at a frequency of $\SI{7}{\hertz}$. This tracking was performed by time-synchronizing the images with the transducer poses and interpolating the recorded poses to the image timestamps. The transducer positions were interpolated linearly, and the robot orientations were interpolated with spherical linear interpolation.

\begin{figure*}[t]%
    \centering
    \includegraphics[width=\linewidth]{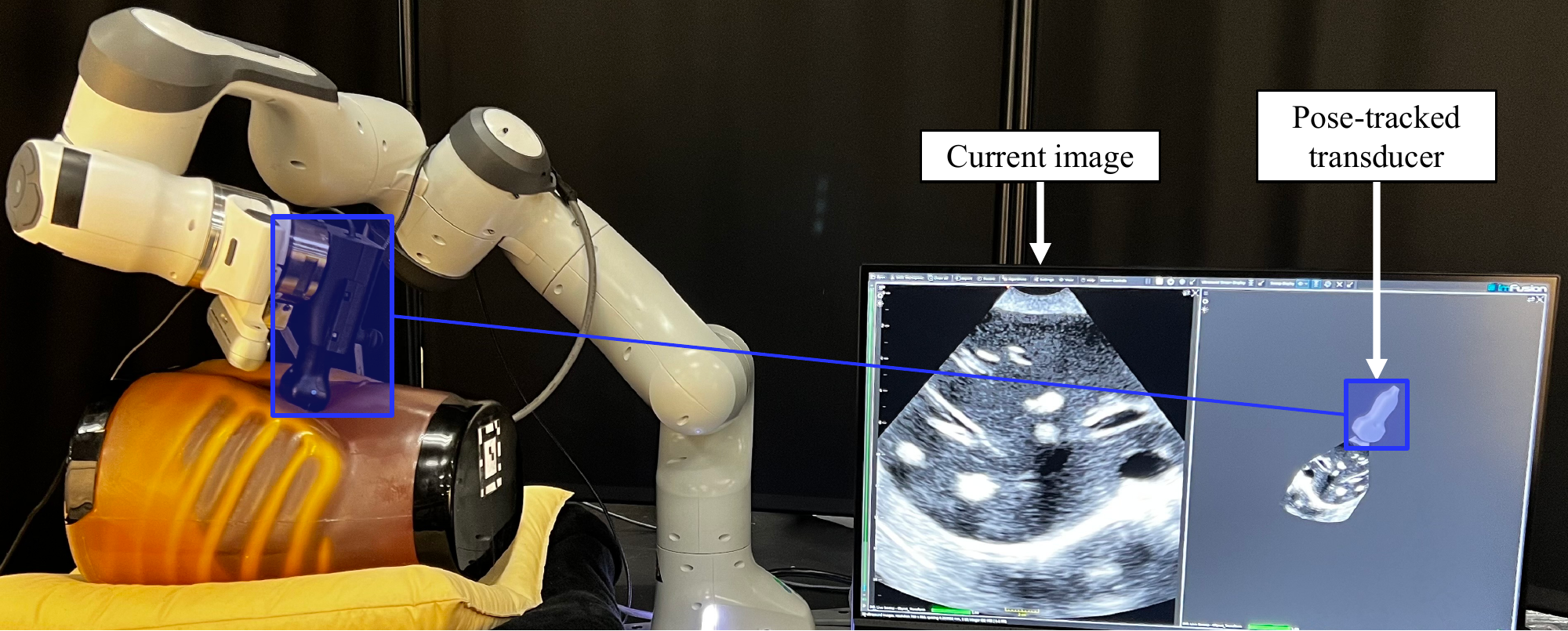}
    \caption{Experimental setup for autonomous robotic ultrasound sweeps. The transducer pose is synchronized with the corresponding ultrasound image to acquire three-dimensional image datasets.}
    \label{fig:3d_setup}
\end{figure*}

The pose-tracked images enabled the acquisition of three-dimensional image datasets, which were compounded into three-dimensional volumes.
Lesion segmentation was performed on the resulting anatomical volumes.~\Cref{fig:3d_setup} shows the experimental setup for the autonomous scan along with ImFusion~\citep{Zettinig2018}, the software used to record the three-dimensional image data and compute the corresponding volumes and lesion segmentations.

\subsubsection{Motion and Force Tracking Performance.}
\begin{figure*}[t]%
    \centering
    \includegraphics[width=\linewidth]{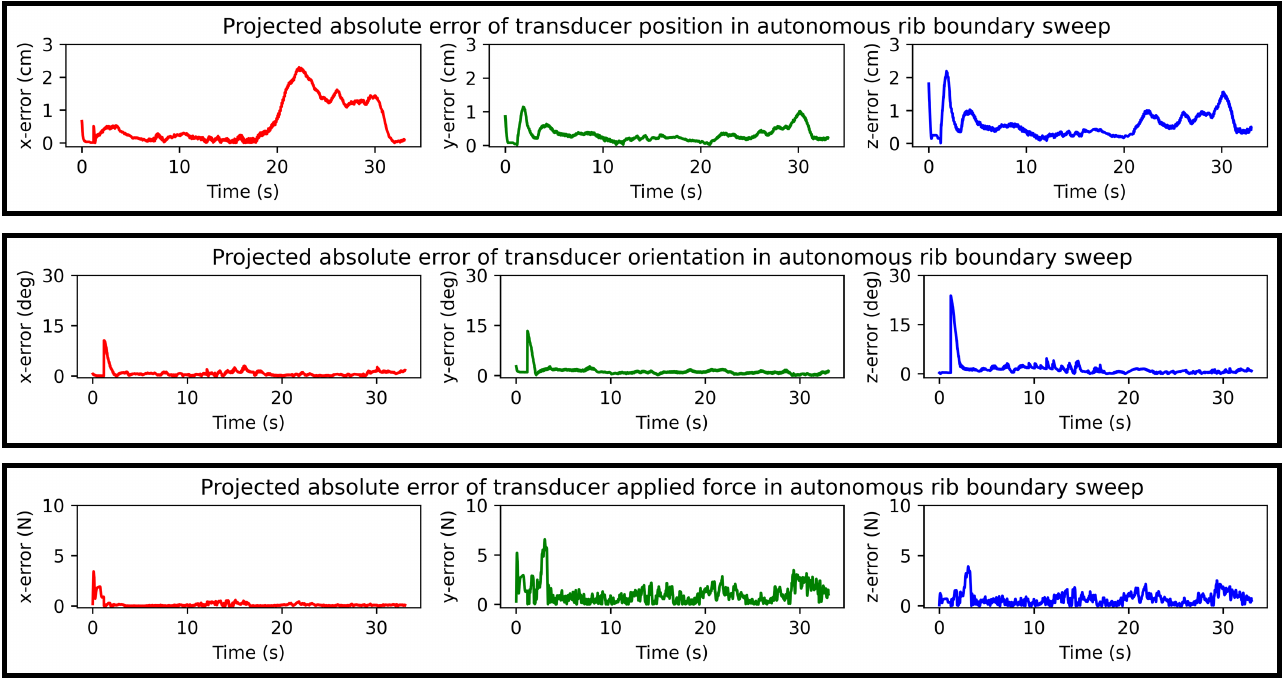}
    \caption{Projected transducer motion and force error during autonomous rib boundary sweep.}
    \label{fig:rib_sweep_projected_errors}
\end{figure*}

\begin{figure*}[t]%
    \centering
    \includegraphics[width=\linewidth]{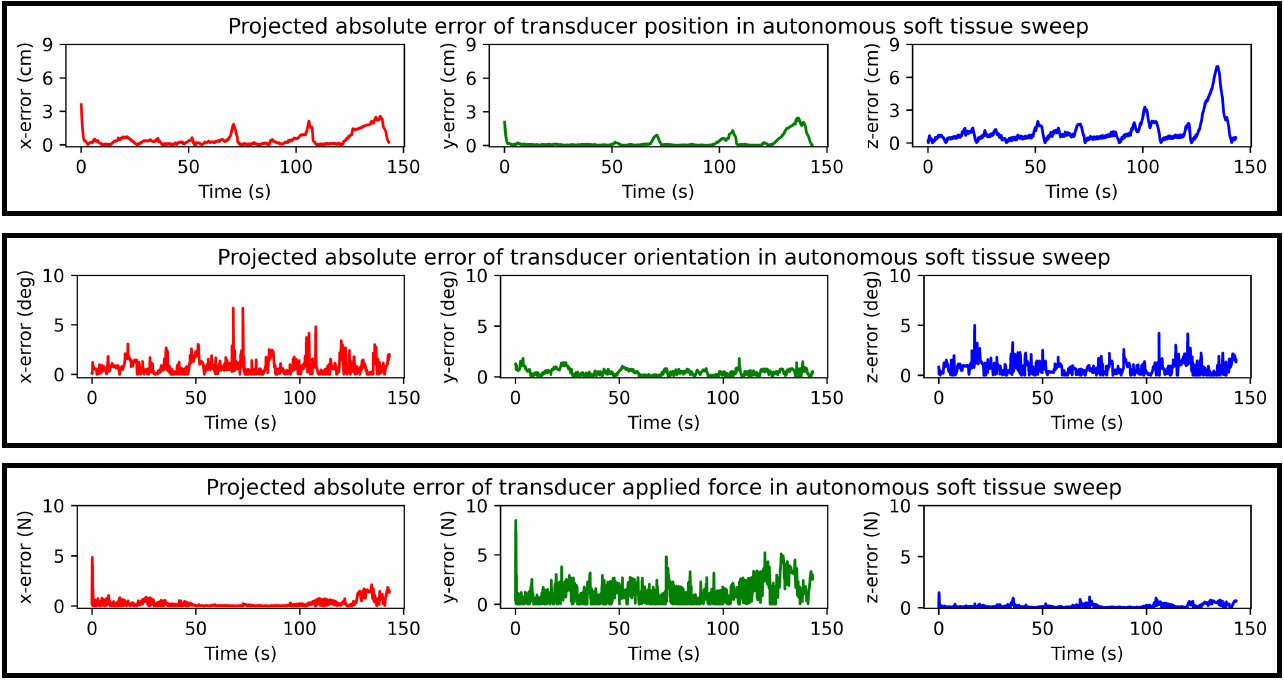}
    \caption{Projected transducer motion and force error during autonomous soft tissue sweep.}
    \label{fig:up_sweep_projected_errors}
\end{figure*}

The error magnitudes between the desired and actual linear and rotational motion, as well as linear applied force, projected into the extracted motion and force spaces are shown in~\Cref{fig:rib_sweep_projected_errors,fig:up_sweep_projected_errors}.

During the rib boundary sweep, the average motion space position error magnitude along the $x$, $y$, and $z$-directions were $\SI{0.62}{\cm}$, $\SI{0.31}{\cm}$, and $\SI{0.57}{\cm}$, respectively. Those same values were $\SI{0.48}{\cm}$, $\SI{0.27}{\cm}$, and $\SI{1.08}{\cm}$ during the soft tissue sweep. The average orientation error magnitude about the $x$, $y$, and $z$-directions were $\SI{0.8}{\deg}$, $\SI{1.2}{\deg}$, and $\SI{1.4}{\deg}$, respectively. Those values were $\SI{0.7}{\deg}$, $\SI{0.4}{\deg}$, and $\SI{0.7}{\deg}$ during the soft tissue sweep. The average force space linear force error magnitude along the projected $x$, $y$, and $z$-directions were $\SI{0.16}{\N}$, $\SI{1.08}{\N}$, and $\SI{0.68}{\N}$, respectively. During the soft tissue sweep, linear force error magnitudes were $\SI{0.26}{\N}$, $\SI{1.22}{\N}$, and $\SI{0.14}{\N}$.

\subsubsection{Volume Reconstruction and Lesion Segmentation.}
\begin{figure*}[t]%
    \centering
    \includegraphics[width=\linewidth]{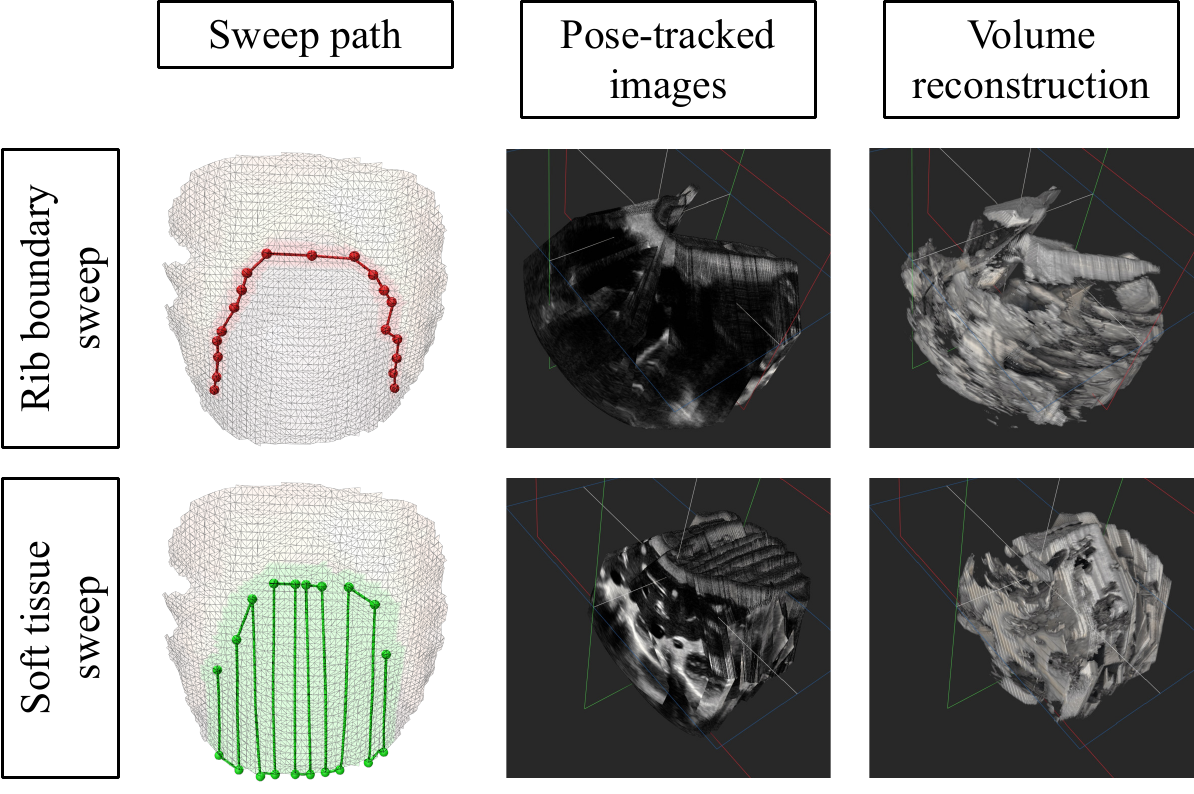}
    \caption{Three-dimensional image datasets and corresponding volumes from the two autonomous ultrasound sweeps.}
    \label{fig:3d_volumes}
\end{figure*}

\begin{figure*}[t]%
    \centering
    \includegraphics[width=\linewidth]{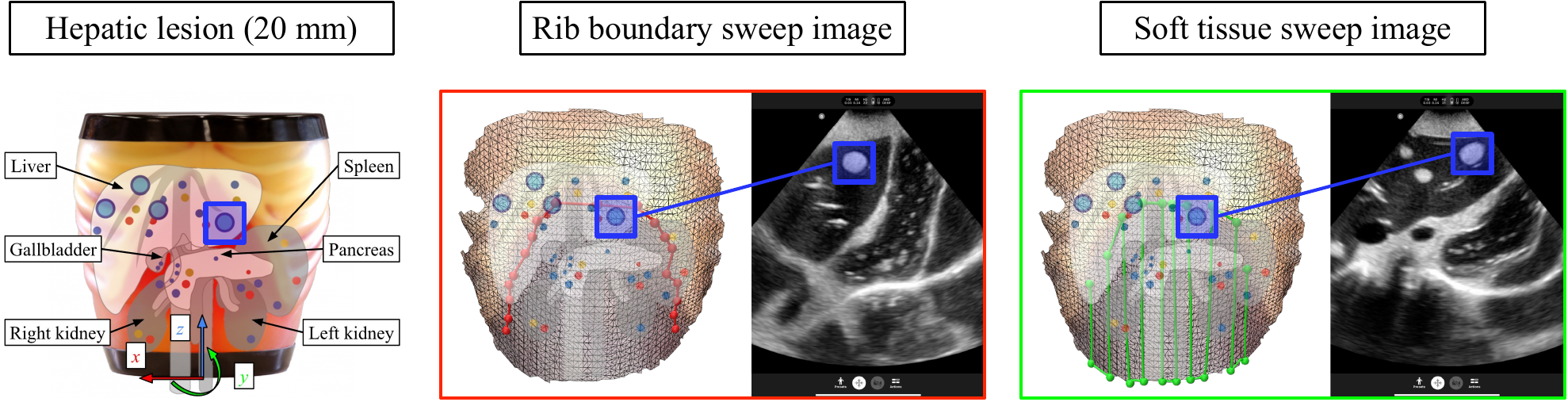}
    \caption{Hepatic lesion highlighted in the phantom model and images taken from the rib boundary and soft tissue sweeps.}
    \label{fig:hepatic_lesion}
\end{figure*}

The pose-tracked images acquired during the rib boundary and soft tissue sweeps are shown in~\Cref{fig:3d_volumes}, along with the volumes computed from each image dataset. An important aspect of performing an ultrasound scan is to properly identify any lesions, and in particular we will focus on a focal echogenic hepatic lesion simulating a metastasis. According to the phantom manufacturer, the lesion has a diameter of $\SI{20}{\milli\meter}$. A reference image of the lesion is provided in~\Cref{fig:hepatic_lesion}, along with ultrasound images of said lesion from the rib boundary and soft tissue sweeps.


\begin{figure*}[t]%
    \centering
    \includegraphics[width=\linewidth]{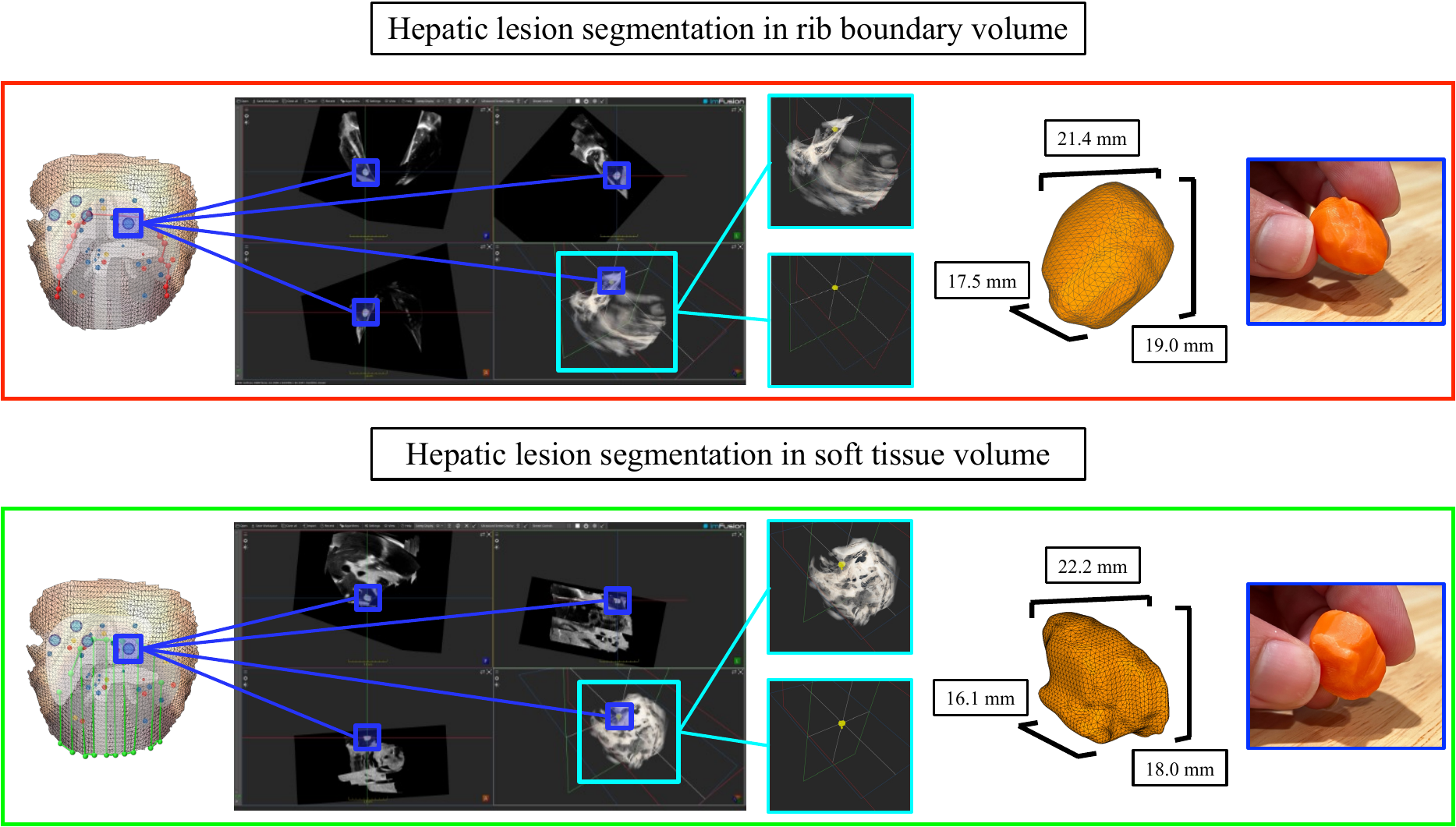}
    \caption{Hepatic lesion segmentation process for the two autonomous ultrasound sweeps. \textbf{Left-to-right}: For each sweep, after the three-dimensional image volume was computed, the hepatic lesion was identified in the three multiplanar reformation views. The lesion was segmented in the volume, exported as a model with specified dimensions, and physically printed.}
    \label{fig:3d_segmentations}
\end{figure*}

The hepatic lesion was identified in the three multiplanar reformation (MPR) views. A coarse segmentation of the lesion was performed by manually classifying the portions of the MPR images internal and external to the lesion. The MPR segmentations were combined to generate a three-dimensional segmentation of the lesion within the computed image volumes. The lesion segmentation was exported and physically printed.~\Cref{fig:3d_segmentations} provides illustrations of the lesion segmentation process in both autonomous sweeps.

The minimum bounding boxes around the lesion segmented from the rib boundary and soft tissue sweeps measure $\SI{7.11}{\cubic\centi\meter}$ and $\SI{6.43}{\cubic\centi\meter}$, respectively. Assuming that the nominal $\SI{20}{\milli\meter}$ lesion diameter corresponds to a volume of $\SI{8}{\cubic\centi\meter}$, the measured volumes are smaller by a factor of $\SI{112}{\percent}$ and $\SI{124}{\percent}$.

%% file: sections/6-conclusion.tex
\section{Conclusion}\label{sec:conclusion}

Our framework and procedure for performing autonomous scans with a robotic ultrasound acquisition system represents a significant step toward generalizing robotic imaging for diverse patients and clinical scenarios. The system integrates stereo vision, touch-based perception, and expert-informed strategies to dynamically adapt to patient-specific anatomy.
We highlight the following results:
\begin{enumerate}
    \item To investigate the scanning strategies used by skilled ultrasound physicians, we developed a framework to record their freehand transducer motion, applied forces, and generated ultrasound images.
    \item We developed a robotic ultrasound acquisition system. This system can either be controlled remotely by an operator or to characteristically replay recorded motion and force data.
    \item Using the recorded narration from the freehand ultrasound scan performed by an expert, we separated the full scan into segments focusing on distinct organs in the transverse and sagittal plans. The scan was optimized to remove redundant portions. The processed data was input to the robot controller to perform a robotic replay scan. Images taken from the most similar points in the freehand and robotically-acquired scans show the same anatomy and pathologies, thus demonstrating functional equivalence between the skilled ultrasound physician and robotic system.
    \item The autonomous capabilities of the robot ultrasound acquisition system were extended. Stereo vision was incorporated to create patient-specific topography maps and initial ultrasound scan paths. Delineation of the rib cage boundary was enabled by linear stiffness measurements at key points. Knowledge of the rib cage boundary was used to refine the vision-based scan paths and enable the human expert strategy of scanning underneath the ribs and normal to the soft tissue.
    \item The robotic ultrasound system enabled three-dimensional image acquisition, a key advantage over freehand scanning. The pose-tracked ultrasound images from the rib boundary and soft tissue sweeps were compounded into three-dimensional volumes. A focal echogenic hepatic lesion simulating a metastasis was segmented from both volumes, further demonstrating the diagnostic capabilities of the robotic system.
\end{enumerate}



While the presented robotic ultrasound system demonstrates significant advancements in autonomous imaging, several avenues exist for further development. One promising direction is the integration of adaptive transducer velocity and pressure modulation to optimize image resolution and minimize noise in real-time. By dynamically adjusting these parameters based on tissue properties and imaging requirements, the system could further enhance diagnostic performance. Improving safety is another critical focus, such as implementing predictive algorithms to identify and avoid potential issues like excessive force on sensitive areas or unintended probe movements. Additionally, this study was conducted using a static phantom, which does not account for the complexities introduced by moving anatomy or a breathing patient. Extending the touch-based perception approach to accommodate dynamic anatomical changes could ensure robust operation in real-world clinical settings. Future work may also explore the incorporation of advanced learning techniques to enable the system to generalize scanning strategies across a wider range of patients and anatomical variations. Furthermore, integrating multimodal sensing, such as thermal imaging or Doppler ultrasound, could expand the system's diagnostic capabilities by providing complementary information about tissue properties and blood flow. These enhancements will pave the way for more versatile and reliable robotic ultrasound systems, capable of addressing a broader spectrum of clinical challenges.

%% file: main.bbl
\begin{thebibliography}{31}
\providecommand{\natexlab}[1]{#1}
\providecommand{\url}[1]{\texttt{#1}}
\providecommand{\urlprefix}{URL }
\expandafter\ifx\csname urlstyle\endcsname\relax
  \providecommand{\doi}[1]{DOI:\discretionary{}{}{}#1}\else
  \providecommand{\doi}{DOI:\discretionary{}{}{}\begingroup \urlstyle{rm}\Url}\fi

\bibitem[{Athitsos et~al.(2008)Athitsos, Potamias, Papapetrou and Kollios}]{Athitsos2008}
Athitsos V, Potamias M, Papapetrou P and Kollios G (2008) Nearest neighbor retrieval using distance-based hashing.
\newblock \emph{Proceedings - International Conference on Data Engineering} : 327--336\doi{10.1109/ICDE.2008.4497441}.

\bibitem[{Bamaarouf et~al.(2024)Bamaarouf, Paccot, Sarry and Chanal}]{Bamaarouf2024}
Bamaarouf M, Paccot F, Sarry L and Chanal H (2024) Development of a robotic ultrasound system to assist ultrasound examination of pregnant women.
\newblock \emph{IEEE Transactions on Medical Robotics and Bionics} 6: 796--805.
\newblock \doi{10.1109/TMRB.2024.3387047}.

\bibitem[{Bharadwaj et~al.(2022)Bharadwaj, Shah, Zhao, Harindranath, George, Krishnan and Arora}]{Bharadwaj2022}
Bharadwaj S, Shah K, Zhao Y, Harindranath A, George A, Krishnan K and Arora M (2022) Semi-blinded freehand 3d ultrasound with novice users from indian institute of science, national institute of advanced studies, cranfield university and st.john's medical college hospital.
\newblock \doi{10.1117/12.2610969}.

\bibitem[{Bi et~al.(2024)Bi, Jiang, Duelmer, Huang and Navab}]{Bi2024}
Bi Y, Jiang Z, Duelmer F, Huang D and Navab N (2024) Machine learning in robotic ultrasound imaging: Challenges and perspectives.
\newblock \emph{Annu. Rev. Control. Robotics Auton. Syst.} 7: 335--357.
\newblock \doi{10.1146/ANNUREV-CONTROL-091523-100042}.

\bibitem[{Chatelain et~al.(2015)Chatelain, Krupa and Navab}]{Chatelain2015}
Chatelain P, Krupa A and Navab N (2015) Optimization of ultrasound image quality via visual servoing.
\newblock \emph{Proceedings - IEEE International Conference on Robotics and Automation} 2015-June: 5997--6002.
\newblock \doi{10.1109/ICRA.2015.7140040}.

\bibitem[{Conti et~al.(2014)Conti, Park and Khatib}]{Conti2014}
Conti F, Park J and Khatib O (2014) Interface design and control strategies for a robot assisted ultrasonic examination system.
\newblock \emph{Springer Tracts in Advanced Robotics} 79: 97--113.
\newblock \doi{10.1007/978-3-642-28572-1_7/COVER}.

\bibitem[{Duffy et~al.(2024)Duffy, Christensen and Ouyang}]{Duffy2024}
Duffy G, Christensen K and Ouyang D (2024) Leveraging 3d echocardiograms to evaluate ai model performance in predicting cardiac function on out-of-distribution data.
\newblock In: \emph{Pacific Symposium on Biocomputing}.
\newblock \doi{10.1142/9789811286421_0004}.

\bibitem[{Fu and Cai(2021)}]{Fu2021}
Fu B and Cai G (2021) Design and calibration of a joint torque sensor for robot compliance control.
\newblock \emph{IEEE Sensors Journal} 21: 21378--21389.
\newblock \doi{10.1109/JSEN.2021.3104351}.

\bibitem[{Fu et~al.(2023)Fu, Lin, Yu, Rodriguez-Andina and Gao}]{Fu2023}
Fu Y, Lin W, Yu X, Rodriguez-Andina JJ and Gao H (2023) Robot-assisted teleoperation ultrasound system based on fusion of augmented reality and predictive force.
\newblock \emph{IEEE Transactions on Industrial Electronics} 70: 7449--7456.
\newblock \doi{10.1109/TIE.2022.3201322}.

\bibitem[{Huang and Zeng(2017)}]{Huang2017}
Huang Q and Zeng Z (2017) A review on real-time 3d ultrasound imaging technology.
\newblock \emph{BioMed Research International} 2017.
\newblock \doi{10.1155/2017/6027029}.

\bibitem[{Jiang et~al.(2023)Jiang, Salcudean and Navab}]{Jiang2023}
Jiang Z, Salcudean SE and Navab N (2023) Robotic ultrasound imaging: State-of-the-art and future perspectives.
\newblock \emph{Medical Image Analysis} 89: 102878.
\newblock \doi{10.1016/J.MEDIA.2023.102878}.

\bibitem[{Jorda et~al.(2022)Jorda, Vulliez and Khatib}]{Jorda2022}
Jorda M, Vulliez M and Khatib O (2022) Local autonomy-based haptic-robot interaction with dual-proxy model.
\newblock \emph{IEEE Transactions on Robotics} \doi{10.1109/TRO.2022.3160053}.

\bibitem[{Kazhdan et~al.(2006)Kazhdan, Bolitho and Hoppe}]{Kazhdan2006}
Kazhdan M, Bolitho M and Hoppe H (2006) Poisson surface reconstruction.
\newblock \emph{Eurographics Symposium on Geometry Processing} .

\bibitem[{Khatib(1987)}]{Khatib1987}
Khatib O (1987) A unified approach for motion and force control of robot manipulators: The operational space formulation.
\newblock \emph{IEEE Journal on Robotics and Automation} 3: 43--53.
\newblock \doi{10.1109/JRA.1987.1087068}.

\bibitem[{Kröger(2010)}]{Kroeger2010}
Kröger T (2010) On-line trajectory generation in robotic systems: Basic concepts for instantaneous reactions to unforeseen (sensor) events.
\newblock \emph{Springer Tracts in Advanced Robotics} 58: 1--230.
\newblock \doi{10.1007/978-3-642-05175-3_1}.

\bibitem[{Kunisch and Volkwein(1999)}]{Kunisch1999}
Kunisch K and Volkwein S (1999) Control of the burgers equation by a reduced-order approach using proper orthogonal decomposition.
\newblock \emph{Journal of Optimization Theory and Applications} 102: 345--371.
\newblock \doi{10.1023/A:1021732508059/METRICS}.

\bibitem[{Kuo et~al.(2023)Kuo, Ma, Deshmukh and Zhang}]{Kuo2023}
Kuo WY, Ma X, Deshmukh D and Zhang HK (2023) Automatic contact force-regulated end-effector using pneumatic actuator for safe robotic ultrasound imaging.
\newblock \emph{International Symposium on Medical Robotics} \doi{10.1109/ISMR57123.2023.10130200}.

\bibitem[{Li et~al.(2021)Li, Xu and Meng}]{Li2021}
Li K, Xu Y and Meng MQ (2021) An overview of systems and techniques for autonomous robotic ultrasound acquisitions.
\newblock \emph{IEEE Transactions on Medical Robotics and Bionics} 3: 510--524.
\newblock \doi{10.1109/TMRB.2021.3072190}.

\bibitem[{Machado et~al.(2018)Machado, Toews, Luo, Unadkat, Essayed, George, Teodoro, Carvalho, Martins, Golland, Pieper, Frisken, Golby and Wells}]{Machado2018}
Machado I, Toews M, Luo J, Unadkat P, Essayed W, George E, Teodoro P, Carvalho H, Martins J, Golland P, Pieper S, Frisken S, Golby A and Wells W (2018) Non-rigid registration of 3d ultrasound for neurosurgery using automatic feature detection and matching.
\newblock \emph{International journal of computer assisted radiology and surgery} 13: 1525.
\newblock \doi{10.1007/S11548-018-1786-7}.

\bibitem[{Piedra et~al.(2024)Piedra, Jeffrey and Khatib}]{Piedra2024}
Piedra A, Jeffrey RB and Khatib O (2024) Robotic ultrasound imaging with haptic guidance and human expert strategies.
\newblock In: \emph{Springer Proceedings in Advanced Robotics}, volume~30. Springer Nature, pp. 53--65.
\newblock \doi{10.1007/978-3-031-63596-0_6}.

\bibitem[{Raina et~al.(2024)Raina, Zhao, Voyles, Wachs, Saha and Chandrashekhara}]{Raina2024}
Raina D, Zhao Z, Voyles R, Wachs J, Saha SK and Chandrashekhara SH (2024) Ultragelbot: Autonomous gel dispenser for robotic ultrasound.
\newblock \emph{ArXiv} : 119--120\doi{10.31256/HSMR2024.60}.

\bibitem[{Salinaro et~al.(2021)Salinaro, McNally, Vissoci, Ellestad, Nelson and Broder}]{Salinaro2021}
Salinaro JR, McNally PJ, Vissoci JRN, Ellestad SC, Nelson B and Broder JS (2021) A prospective blinded comparison of second trimester fetal measurements by expert and novice readers using low-cost novice-acquired 3d volumetric ultrasound.
\newblock \emph{Journal of Maternal-Fetal and Neonatal Medicine} 34.
\newblock \doi{10.1080/14767058.2019.1649390}.

\bibitem[{Santos and Cortesao(2018)}]{Santos2018}
Santos L and Cortesao R (2018) Computed-torque control for robotic-assisted tele-echography based on perceived stiffness estimation.
\newblock \emph{IEEE Transactions on Automation Science and Engineering} 15: 1337--1354.
\newblock \doi{10.1109/TASE.2018.2790900}.

\bibitem[{Senthilkumar and Ezhilarasi(2016)}]{Senthilkumar2016}
Senthilkumar V and Ezhilarasi M (2016) An efficient multi-rvm classification-based ultrasound lung image retrieval approach.
\newblock \emph{International Journal of Biomedical Engineering and Technology} 22: 189--215.
\newblock \doi{10.1504/IJBET.2016.079485}.

\bibitem[{Sharp and Crane(2020)}]{Sharp2020}
Sharp N and Crane K (2020) You can find geodesic paths in triangle meshes by just flipping edges.
\newblock \emph{ACM Trans. Graph} 39: 15.
\newblock \doi{10.1145/3414685.3417839}.

\bibitem[{Shyam et~al.(2023)Shyam, Purayath, Keerthivasan, Akash, Govindaraju, Lakshmanan and Sivaprakasam}]{Shyam2023}
Shyam A, Purayath A, Keerthivasan S, Akash SM, Govindaraju A, Lakshmanan M and Sivaprakasam M (2023) Immersive virtual reality platform for robot-assisted antenatal ultrasound scanning.
\newblock \emph{IEEE International Symposium on Robot and Human Interactive Communication} : 1600--1605\doi{10.1109/RO-MAN57019.2023.10309266}.

\bibitem[{Si et~al.(2024)Si, Wang and Yang}]{Si2024}
Si W, Wang N and Yang C (2024) Design and quantitative assessment of teleoperation-based human–robot collaboration method for robot-assisted sonography.
\newblock \emph{IEEE Transactions on Automation Science and Engineering} \doi{10.1109/TASE.2024.3350524}.

\bibitem[{Tang et~al.(2024)Tang, Wang, Luo, Jiang, Nian, Qi, Sang and Gan}]{Tang2024}
Tang X, Wang H, Luo J, Jiang J, Nian F, Qi L, Sang L and Gan Z (2024) Autonomous ultrasound scanning robotic system based on human posture recognition and image servo control: an application for cardiac imaging.
\newblock \emph{Frontiers in robotics and AI} 11.
\newblock \doi{10.3389/FROBT.2024.1383732}.

\bibitem[{Valente et~al.(2022)Valente, Morais, Torres, Oliveira, Gomes-Fonseca, Buschle, Fritz, Correia-Pinto, Lima and Vilaça}]{Valente2022}
Valente S, Morais P, Torres HR, Oliveira B, Gomes-Fonseca J, Buschle LR, Fritz A, Correia-Pinto J, Lima E and Vilaça JL (2022) A deep learning method for kidney segmentation in 2d ultrasound images.
\newblock In: \emph{Proceedings of the Annual International Conference of the IEEE Engineering in Medicine and Biology Society, EMBS}, volume 2022-July.
\newblock \doi{10.1109/EMBC48229.2022.9871748}.

\bibitem[{von Haxthausen et~al.(2021)von Haxthausen, Böttger, Wulff, Hagenah, García-Vázquez and Ipsen}]{Haxthausen2021}
von Haxthausen F, Böttger S, Wulff D, Hagenah J, García-Vázquez V and Ipsen S (2021) Medical robotics for ultrasound imaging: Current systems and future trends.
\newblock \emph{Current robotics reports} 2: 55--71.
\newblock \doi{10.1007/S43154-020-00037-Y}.

\bibitem[{Zettinig et~al.(2018)Zettinig, Salehi, Prevost and Wein}]{Zettinig2018}
Zettinig O, Salehi M, Prevost R and Wein W (2018) Recent advances in point-of-care ultrasound using the imfusion suite for real-time image analysis.
\newblock \emph{Lecture Notes in Computer Science (including subseries Lecture Notes in Artificial Intelligence and Lecture Notes in Bioinformatics)} 11042 LNCS: 47--55.
\newblock \doi{10.1007/978-3-030-01045-4_6/FIGURES/4}.

\end{thebibliography}
